\documentclass{article}

\PassOptionsToPackage{numbers,compress}{natbib}
\usepackage[preprint]{neurips_2026}

\usepackage[utf8]{inputenc} % allow utf-8 input
\usepackage[T1]{fontenc}    % use 8-bit T1 fonts
\usepackage{hyperref}       % hyperlinks
\usepackage{url}            % simple URL typesetting
\usepackage{booktabs}       % professional-quality tables
\usepackage{amsfonts}   
\usepackage{enumitem}% blackboard math symbols
\usepackage{nicefrac}       % compact symbols for 1/2, etc.
\usepackage{microtype}      % microtypography
\usepackage[table]{xcolor}  % colors and table row shading
\usepackage{amsmath}
\usepackage{graphicx}
\usepackage{algorithm}
\usepackage{algorithmic}
\usepackage{multirow}
\usepackage{subcaption}
\usepackage{xspace}
\usepackage{listings}
\usepackage[listings,skins,breakable]{tcolorbox}
\usepackage{titletoc} % per-section (appendix-only) table of contents
\usepackage{tikz}
\usepackage{pifont}

\newcommand{\ours}{VeriGraph\xspace}
\definecolor{TableGroup}{HTML}{F3F5F7}
\definecolor{OursRow}{HTML}{EEF4FF}
\definecolor{TraceRed}{HTML}{D64545}
\definecolor{TraceAmber}{HTML}{E8A33D}
\definecolor{TraceGreen}{HTML}{2E8B57}
\definecolor{PromptFrame}{HTML}{78848F}
\definecolor{PromptBack}{HTML}{F7F9FB}
\definecolor{PromptTitleBack}{HTML}{46515D}
\hypersetup{pdfborder={0 0 0}}
\definecolor{AppendixTOCSection}{HTML}{1F4E79}
\definecolor{AppendixTOCSubsection}{HTML}{2F75B5}
\definecolor{AppendixTOCSubsubsection}{HTML}{5B9BD5}
\newcommand{\setupappendixtoc}{%
    \titlecontents{section}
        [2.6em]
        {\addvspace{0.35em}\small\bfseries\color{AppendixTOCSection}}
        {\contentslabel{2.3em}}
        {}
        {\titlerule*[0.45pc]{.}\contentspage}
    \titlecontents{subsection}
        [3.9em]
        {\small\color{AppendixTOCSubsection}}
        {\contentslabel{2.8em}}
        {}
        {\titlerule*[0.45pc]{.}\contentspage}
    \titlecontents{subsubsection}
        [6.2em]
        {\footnotesize\color{AppendixTOCSubsubsection}}
        {\contentslabel{3.5em}}
        {}
        {\titlerule*[0.45pc]{.}\contentspage}%
}
\DeclareRobustCommand{\tracenone}{\textcolor{TraceRed}{\ding{55}}}

\DeclareRobustCommand{\tracefull}{\textcolor{TraceGreen}{\ding{51}}}
\newcommand{\best}[1]{\textbf{#1}}
\newcommand{\second}[1]{\underline{#1}}
\newcommand{\pending}{--}
\newtcblisting[auto counter]{promptbox}[3][]{%
    enhanced jigsaw,
    breakable,
    pad at break*=1mm,
    colframe=PromptFrame,
    colback=PromptBack,
    colbacktitle=PromptTitleBack,
    coltitle=white,
    coltext=black,
    fonttitle=\bfseries\footnotesize,
    title={Listing~\thetcbcounter: #2},
    label={#3},
    boxrule=0.8pt,
    arc=1.2mm,
    width=\linewidth,
    left=6pt,
    right=6pt,
    top=5pt,
    bottom=5pt,
    listing only,
    listing engine=listings,
    listing options={
        basicstyle=\ttfamily\scriptsize,
        breaklines=true,
        columns=fullflexible,
        keepspaces=true,
        showstringspaces=false,
        frame=none,
        tabsize=2
    },
    #1
}

\title{VeriGraph: Towards Verifiable Data-Analytic Agents}

\author{
  \textbf{Jiajie Jin}\textsuperscript{1}\thanks{Equal contribution.},
  \textbf{Zhao Yang}\textsuperscript{1}\footnotemark[1],
  \textbf{Wenle Liao}\textsuperscript{1},
  \textbf{Yuyang Hu}\textsuperscript{1},\\
  \textbf{Guanting Dong}\textsuperscript{1},
  \textbf{Xiaoxi Li}\textsuperscript{1},
  \textbf{Yutao Zhu}\textsuperscript{1},
  \textbf{Zhicheng Dou}\textsuperscript{1}\thanks{Corresponding author.} \\
  \textsuperscript{1}Gaoling School of Artificial Intelligence, Renmin University of China \\
  \texttt{\{jinjiajie, dou\}@ruc.edu.cn}
}

\begin{document}

\maketitle

\begin{abstract}
LLM-based agents have demonstrated strong capabilities in data-intensive analytical tasks, yet their outputs are rarely \emph{verifiable}: a reliance on linear text trajectories makes their reasoning difficult to audit. In particular, deterministic computations over raw data and semantic deductions over natural-language claims are often entangled in an unstructured stream, leaving numerical conclusions hard to reproduce and qualitative judgments hard to inspect. To address this, we propose \ours, a traceable neuro-symbolic reasoning framework that enables agents to construct an explicit heterogeneous evidence directed acyclic graph (DAG) during execution. \ours introduces three evidence-expansion primitives, namely computational, grounding, and derivational expansion, to connect raw data, interpreter variables, computed results, and natural-language claims in a unified graph. Under this formulation, structural traceability is reduced to graph reachability from raw data sources to terminal claims, while semantic support is measured by claim-level evidence evaluation. To improve graph construction, we further design a graph-based policy optimization strategy with a composite reward that jointly supervises answer correctness, computational integrity, and derivational coherence. Experiments on four benchmarks show that \ours-8B achieves the highest overall score among all baselines. More importantly, \ours produces auditable evidence graphs with substantially stronger claim grounding, achieving a 87.61\% Grounding Rate under our claim-level evidence support evaluation. These results suggest that explicit evidence-graph construction is a promising path toward \emph{verifiable data-analytic agents}. Our code is available at \url{https://github.com/ignorejjj/VeriGraph}.
\end{abstract}

\section{Introduction}

Large Language Model (LLM)-based agents~\cite{llm-survey, agentsurvey} have recently demonstrated strong capabilities in tool use~\cite{react, wang2024codeact, ToolLLM}, code generation~\cite{code_generation_survey, swebench}, and multi-step reasoning~\cite{jin2025searchr1, searcho1}. A particularly impactful application of these capabilities is the \emph{data-analytic agent}, which couples an LLM with a code interpreter to tackle data-intensive analytical tasks such as financial analysis~\cite{finsight, financial_report_chunking} and data science~\cite{datamind, datacopilot}. In such tasks, however, trustworthy generation requires more than final-answer accuracy~\cite{finsight,deepanalyze,Establish_Trustworthiness}: outputs must be \emph{verifiable}, i.e., users must be able to check how each conclusion is obtained, especially when the answer depends on external data, numerical computation, and multi-step interpretation~\cite{trustworthy_survey, trustagent}. For instance, a claim like ``Q3 revenue grew $12.3\%$ year-over-year'' is trustworthy only when the number is computed from the underlying transaction tables rather than asserted by the model, and that computation is exposed for the reader to verify. This entails two evidence requirements. Quantitative claims must be reproducible from raw data through deterministic computations, and qualitative judgments must be grounded in inspectable reasoning chains.

\begin{figure*}[!t]
    \centering
    \includegraphics[width=\linewidth]{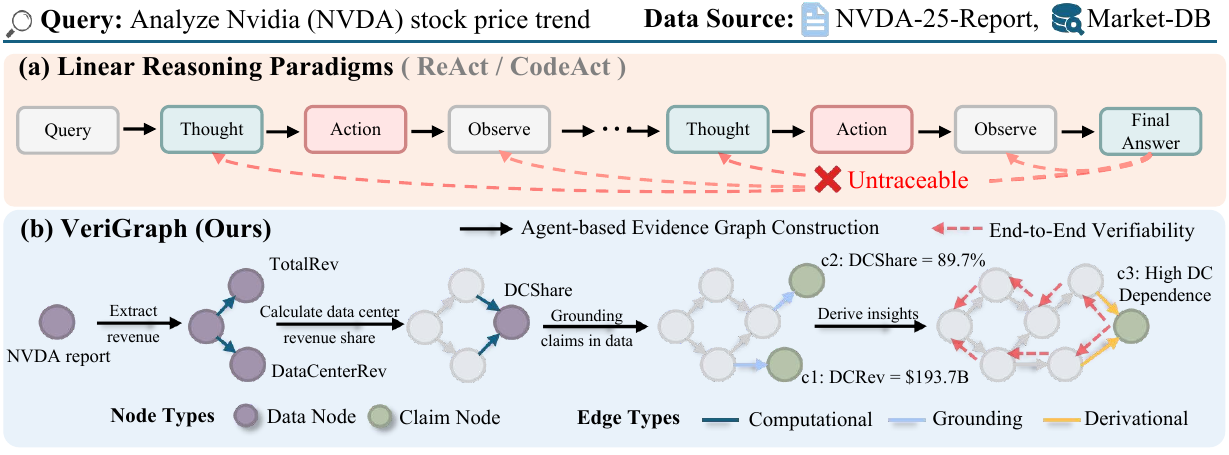}
    \caption{
    Comparison between our proposed \ours and linear reasoning paradigms.
    }
    \label{fig:intro_fig}
    \vspace{-1.4em}
\end{figure*}

Current agent paradigms provide little support for this requirement. As illustrated in Figure~\ref{fig:intro_fig}, prevailing frameworks~\cite{react, wang2024codeact} solve problems through a linear trajectory of thought--action--observation steps and ultimately expose only a final answer. This linear transcript entangles two forms of evidence that should be tracked separately. First, intermediate computational artifacts (e.g., interpreter variables) appear only as transient observations, so numerical claims lose their programmatically recoverable provenance~\cite{datainterpreter, deepanalyze}. Second, the semantic steps that transform computed values into higher-level judgments remain in free-form text, making supported reasoning difficult to distinguish from confabulation~\cite{survey_hallu_nlg, factool, alce, EvaluatingVerifiability}. Recent data-agent work~\cite{datamind, deepanalyze, tabler1} improves end-task accuracy within this paradigm, but still leaves evidence construction outside the learning objective.

Our key observation is that data-intensive reasoning naturally spans two coupled spaces: a \emph{deterministic code space}, where raw data undergo numerical transformations, and a \emph{semantic reasoning space}, where computed results are interpreted and synthesized into higher-level judgments. The provenance crossing these spaces is better represented as a DAG than as a linear transcript. Motivated by this view, we propose VeriGraph (\textbf{Veri}fiable Evidence \textbf{Graph}), a traceable neuro-symbolic reasoning framework that reformulates the agent's objective from emitting an unstructured text stream to incrementally constructing an explicit \emph{heterogeneous evidence DAG}. Its \emph{data nodes} preserve executable provenance over interpreter variables and computed results, while its \emph{claim nodes} expose semantic derivations among natural-language facts and judgments.

To construct the graph during execution, \ours introduces three reasoning primitives aligned with the graph's expansion modes: \emph{computational expansion}, which automatically traces variable dependencies in executed code; \emph{grounding expansion}, which anchors a computed value as an atomic claim with deterministic, re-executable provenance; and \emph{derivational expansion}, which derives a new claim from established premises with an explicit justification. These primitives are embedded directly in the agent's code action space, allowing computation and evidence construction to proceed in one interaction loop. Under this formulation, traceability reduces to graph reachability (\S\ref{sec:graph_def}): a conclusion is verifiable if and only if every constituent claim can be traced backward through the graph to raw data sources. Rather than claiming to completely eliminate hallucinations, \ours makes the reasoning topology explicit so that failures can be localized to the precise unsupported computation, grounding, or derivation step.

Training then focuses on making graph construction reliable rather than merely imitating trajectories. We use synthesized graph-augmented trajectories only as a cold-start stage, teaching the model the basic syntax of the expansion primitives before policy learning. The central challenge is graph-level credit assignment: outcome-only rewards judge the final answer but ignore whether it is supported by connected computations and justified derivations. We therefore introduce \emph{graph-based policy optimization}, whose composite reward mirrors the graph's layered architecture and decomposes credit across answer correctness, computational integrity, and derivational coherence (\S\ref{sec:training}).

We evaluate \ours on four data-intensive benchmarks, including TableBench~\cite{tablellm}, InfiAgent-DABench~\cite{infiagentdabench}, DSBench~\cite{dsbench}, and DAB-Step Research~\cite{dabstep}. \ours-8B achieves the highest Overall score among evaluated baselines while producing explicit evidence graphs, suggesting the effectiveness of structured evidence construction for data-intensive reasoning. We evaluate verifiability on a second axis with \emph{Grounding Rate (GR)}, which decomposes each answer into atomic claims and measures what fraction can be recovered from the evidence artifact exposed by the method.

The core contributions of this paper are summarized as follows:
\begin{itemize}[leftmargin=1em]
    \item \textbf{A traceable neuro-symbolic reasoning framework with an explicit evidence graph.} We propose \ours, which externalizes an LLM agent's implicit reasoning into an executable heterogeneous evidence DAG via computational, grounding, and derivational expansion primitives, enabling conclusions to be traced back to raw data and deterministic computations.
    \item \textbf{Graph-based policy optimization for auditable reasoning.} We identify graph-level credit assignment as the key training challenge and design a composite reward whose terms mirror the graph's layered architecture, jointly supervising answer correctness, computational integrity, and semantic coherence of derivational edges.
    \item \textbf{Comprehensive experimental validation.} On TableBench, DSBench, InfiAgent-DABench, and DAB-Step Research, our 8B \ours achieves the highest Overall score among all baselines. We further show through grounding analysis and ablations that both graph-structured primitives and graph-aware rewards are essential to traceability and accuracy gains.
\end{itemize}

\section{Related Work}

\paragraph{LLM Agents for Data-Intensive Reasoning.}
LLM agents equipped with code interpreters or SQL engines are widely used for data-intensive analysis~\cite{react,wang2024codeact,sqlr1}. Existing work improves them by redesigning agent pipelines~\cite{dsagent,autokaggle,datacopilot,datainterpreter} or scaling task-specific training, from tabular fine-tuning~\cite{tablellm,tabler1,tablegpt2} to trajectory synthesis~\cite{datamind,deepanalyze} on various benchmarks~\cite{tablellm,dsbench}. However, these systems largely optimize the accuracy of the final-answer on flat trajectories~\cite{react,wang2024codeact,datamind,deepanalyze}, leaving implicit the links between code variables and supported claims. Consequently, even strong data agents~\cite{datainterpreter,deepanalyze} remain difficult to audit. \ours addresses this traceability gap by constructing a DAG of executable evidence rather than a linear transcript.

\paragraph{Verifiable Generation and Structural Reasoning.}
Trustworthy-generation work mainly pursues textual attribution and post-hoc verification, including citation or evidence grounding~\cite{alce,EvaluatingVerifiability,CEG} and retrieval mechanisms~\cite{self-ask,self-rag,SelfCheckGPT,Chainofverification}. These methods improve semantic support, but typically treat evidence as text spans rather than deterministic computations with variable provenance. Structural reasoning systems compile problems into programs or symbolic queries~\cite{pal,pot,structgpt,querycompiler}, while verifier-based methods add stronger checks~\cite{factool,fmagent}. Closest to our motivation, recent graph-based frameworks evaluate tool-agent trajectories beyond final-answer matching~\cite{trace2025}, verify reasoning through DAG node blocks~\cite{gov2025}, or reinforce medical reasoning with critical evidence graphs~\cite{medceg2025}. These works contextualize the value of graph structure, but mostly use graphs to evaluate or reward reasoning paths in general tool-use or domain-specific settings. \ours instead makes a heterogeneous evidence DAG the online action interface for data agents, coupling executable code provenance with grounded and derived claims.

\paragraph{Reinforcement Learning for LLM Agents.}
Reinforcement learning has become a standard recipe for training LLM agents~\cite{rlhf,PPO,grpo}, driving progress in mathematical reasoning~\cite{deepseek-r1,deepseek-math}, search-augmented QA~\cite{jin2025searchr1,song2025r1searcher,Li2025WebThinker}, and coding~\cite{start-r1,qian2025toolrl}. Data-analytic agents follow the same recipe, optimizing GRPO-style objectives against final-answer correctness over flat trajectories~\cite{datamind,deepanalyze,tabler1,PPO,deepseek-r1}. However, outcome rewards cannot distinguish a correct answer supported by valid evidence from one reached through unsupported claims. Process rewards such as execution correctness~\cite{deepanalyze} or tool-use efficiency~\cite{stepsearch} help supervise actions, but still ignore graph-level provenance. \ours lifts reward design to the evidence graph, rewarding answer correctness together with raw-data connectivity and local derivation validity.

\section{Methodology}
\label{sec:method}

\subsection{Problem Formulation}
\label{sec:problem}

We consider data-intensive analytical tasks specified by a tuple $(q, \mathcal{F}, d)$, where $q$ is a user query, $\mathcal{F} = \{f_1, \dots, f_n\}$ is a set of heterogeneous data sources (e.g., CSV tables or databases), and $d$ denotes task-specific metadata. Given a ground-truth answer $a^*$, the goal is to produce a response $o$ that is not only accurate, but also auditable: every factual or numerical claim in $o$ should be supported by evidence ultimately originating from $\mathcal{F}$.

Standard data agents instantiate this task as a ReAct-style interaction~\cite{react,wang2024codeact} with a stateful execution environment $\mathcal{E}$ (e.g., a Python interpreter). At step $t$, the agent emits a natural-language \textit{Thought} $\tau_t$ and a code \textit{Action} $\alpha_t$, then receives an \textit{Observation} $z_t = \mathcal{E}(\alpha_t)$. The interaction history to step $t$ is:
\begin{equation}
\label{eq:trajectory}
    h_t = \bigl(q, \mathcal{F}, d,\; \tau_0, \alpha_0, z_0,\; \dots,\; \tau_{t-1}, \alpha_{t-1}, z_{t-1}\bigr),
\end{equation}
and the agent samples $(\tau_t, \alpha_t) \sim \pi_\theta(\cdot \mid h_t)$ until termination or a maximum of $T$ rounds. We retain this interaction interface, but reinterpret the rollout as incrementally constructing an evidence graph $\mathcal{G}$. A response is verifiable only when every claim in $o$ is traceable to the raw sources in $\mathcal{F}$.

\begin{figure*}[!t]
    \centering
    \includegraphics[width=\linewidth]{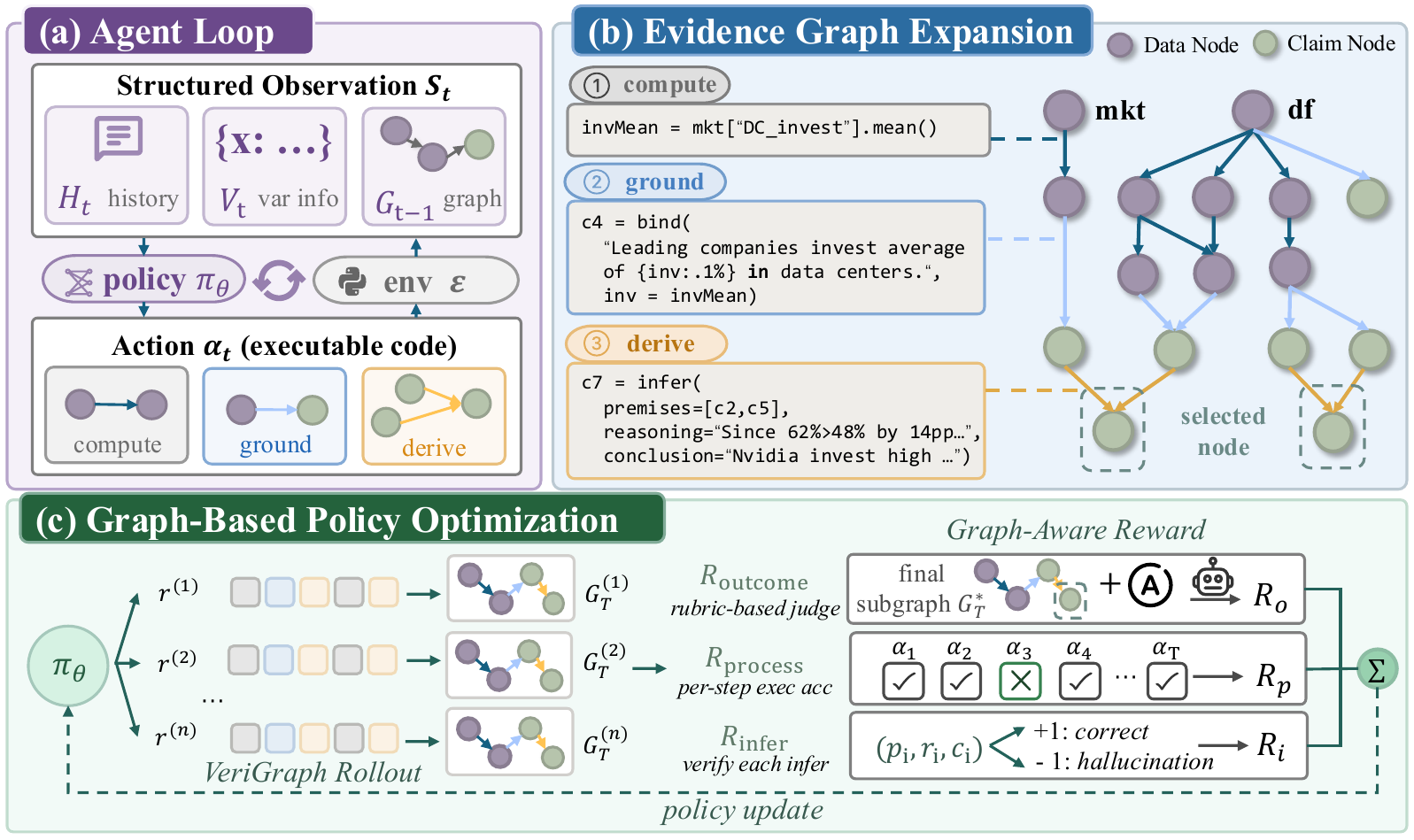}
    \caption{
    Overview of the \ours framework: The agent iteratively generates code to construct a heterogeneous evidence DAG, optimized via a graph-aware composite reward.
    }
    \label{fig:main_fig}
\end{figure*}

\subsection{Overview of VeriGraph}
\label{sec:graph_def}

% To address the limitations of linear agent paradigms, we re-examine the inherent structure of data-driven reasoning. We observe that in data-intensive scenarios, the derivation of a final conclusion is fundamentally a hybrid process: it interleaves deterministic, data-level computations with semantic, natural-language deductions. As the agent continuously bridges these two distinct reasoning spaces, the resulting provenance structure inherently transcends a simple linear sequence, naturally forming a Directed Acyclic Graph (DAG).

% Motivated by this observation, we propose VeriGraph, a traceable neuro-symbolic reasoning framework. VeriGraph reformulates the agent's objective from generating an unstructured text stream to incrementally constructing an explicit \textbf{Heterogeneous Evidence Graph} $\mathcal{G}$. This graph serves as a bridge between a \emph{deterministic code space} (where raw data undergoes numerical transformations) and a \emph{semantic reasoning space} (where computed results are interpreted and synthesized into higher-level judgments). During interaction, the agent expands $\mathcal{G}$ through three topological operations that create data nodes, claim nodes, and directed edges capturing their provenance relations (\S\ref{sec:primitives}). Under this formulation, traceability is rigorously reduced to graph reachability: a conclusion is verifiable if and only if it can be traced back to the raw data through $\mathcal{G}$. 

As illustrated in Figure~\ref{fig:main_fig}(a)--(b), \ours couples a standard agent loop with explicit graph maintenance: each action observes the current history, namespace, and partial graph, then expands the evidence DAG while solving the task. We instantiate this evolving structure as a heterogeneous DAG and formalize auditing as tracing terminal claims back to raw data.

% \paragraph{Graph Structure.}
We define the evidence graph as a DAG $\mathcal{G} = (\mathcal{V}, \mathcal{E})$ whose vertex and edge sets are each partitioned into heterogeneous types:
\begin{equation}
\label{eq:graph}
\mathcal{V} = \mathcal{V}_{\mathrm{data}} \cup \mathcal{V}_{\mathrm{claim}}, \qquad \mathcal{E} = \mathcal{E}_{\mathrm{comp}} \cup \mathcal{E}_{\mathrm{ground}} \cup \mathcal{E}_{\mathrm{derive}}.
\end{equation}
Here, $\mathcal{V}_{\mathrm{data}}$ contains raw sources and intermediate computational artifacts, while $\mathcal{V}_{\mathrm{claim}}$ contains natural-language claims. The three edge types capture the provenance relations used in construction: $\mathcal{E}_{\mathrm{comp}} \subseteq \mathcal{V}_{\mathrm{data}} \times \mathcal{V}_{\mathrm{data}}$ records computational dependencies, $\mathcal{E}_{\mathrm{ground}} \subseteq \mathcal{V}_{\mathrm{data}} \times \mathcal{V}_{\mathrm{claim}}$ grounds artifacts into atomic claims, and $\mathcal{E}_{\mathrm{derive}} \subseteq \mathcal{V}_{\mathrm{claim}} \times \mathcal{V}_{\mathrm{claim}}$ records semantic derivations.

% \paragraph{Traceability as Graph Reachability.}
% Let $\mathcal{V}_{\mathrm{raw}} \subset \mathcal{V}_{\mathrm{data}}$ denote the root nodes corresponding to the raw data sources $\mathcal{F}$, and write $u \leadsto_{\mathcal{G}} v$ if there exists a directed path from $u$ to $v$ in $\mathcal{G}$.

% A claim node $v \in \mathcal{V}_{\mathrm{claim}}$ is \textbf{traceable} if there exists a source node $u \in \mathcal{V}_{\mathrm{raw}}$ and a directed path $P = (v_0, v_1, \dots, v_k)$ in $\mathcal{G}$ satisfying:
% \begin{equation}
% \label{eq:traceable}
% v_0 = u \in \mathcal{V}_{\mathrm{raw}},\quad v_k = v,\quad \forall\, i \in \{0,\dots,k{-}1\}:\;(v_i, v_{i+1}) \in \mathcal{E}.
% \end{equation}
% The response $o$ is \textbf{verifiable} iff every constituent claim satisfies~\eqref{eq:traceable}.

% main figure
% (a) Evidence Graph node and edge types
% (b) A complete agent reasoning example (with bind/infer invocation process)
% (c) Training pipeline (SFT -> RL)

\subsection{Evidence Graph Construction}
\label{sec:primitives}
  
Figure~\ref{fig:main_fig}(b) illustrates one graph-expansion step. At each step $t$, the agent conditions on a structured observation $S_t = (H_t, \mathcal{V}_t, \mathcal{G}_{t-1})$, where $H_t$ is compressed recent history, $\mathcal{V}_t$ is the interpreter namespace, and $\mathcal{G}_{t-1}$ is the current graph, and produces an action that simultaneously advances the task and extends the graph:
\begin{equation}
\label{eq:graph_update}
(\tau_t, \alpha_t) \sim \pi_\theta(\cdot \mid S_t), \quad \mathcal{G}_{t} = \mathcal{G}_{t-1} \;\cup\; \underbrace{\Delta\mathcal{G}_t^{\mathrm{comp}}}_{\text{auto-traced}} \;\cup\; \underbrace{\Delta\mathcal{G}_t^{\mathrm{ground}} \;\cup\; \Delta\mathcal{G}_t^{\mathrm{derive}}}_{\text{agent-invoked}},
\end{equation}
where $\Delta\mathcal{G}_t^{\mathrm{comp}}$ is extracted automatically from execution, while $\Delta\mathcal{G}_t^{\mathrm{ground}}$ and $\Delta\mathcal{G}_t^{\mathrm{derive}}$ are created by agent-invoked primitives embedded in $\alpha_t$: computational expansion tracks how artifacts are produced, \texttt{bind} states what an artifact means, and \texttt{infer} records how the agent reasons over those meanings. Full rollout pseudocode is provided in Appendix~\ref{app:verigraph-algorithm}.

\paragraph{Computational Expansion.}
Executing code automatically expands the code space. Let $\mathrm{NewVars}(\mathcal{E}, \alpha_t)$ denote the set of variables created or modified by action $\alpha_t$, and $\mathrm{Deps}(v, \alpha_t)$ the variables read to compute $v$. The computational subgraph update is:
\begin{equation}
\label{eq:comp_expansion}
\Delta\mathcal{G}_t^{\mathrm{comp}} = \Bigl(\;\mathrm{NewVars}(\mathcal{E}, \alpha_t),\;\; \bigl\{(u, v) \mid v \in \mathrm{NewVars}(\mathcal{E}, \alpha_t),\; u \in \mathrm{Deps}(v, \alpha_t)\bigr\}\;\Bigr).
\end{equation}
In practice, $\mathrm{NewVars}$ comes from pre/post namespace snapshots and $\mathrm{Deps}$ from a static AST walk over $\alpha_t$ (no \texttt{sys.settrace}; Appendix~\ref{app:verigraph-graph}).

\paragraph{Grounding Expansion.}
The \texttt{bind} primitive externalizes the semantic interpretation of an executable artifact. It grounds a runtime variable as an atomic claim whose content is expressed in natural language. Given a data node $v_d \in \mathcal{V}_{\mathrm{data}}$ and a natural-language description $l$ of its value:
\begin{equation}
\label{eq:ground_expansion}
v_c = \texttt{bind}\bigl(v_d,\; l\bigr), \quad \Delta\mathcal{G}_t^{\mathrm{ground}} = \bigl(\{v_c\},\; \{(v_d, v_c)\}\bigr),
\end{equation}
where $v_c \in \mathcal{V}_{\mathrm{claim}}$ is a new atomic claim node. Because \texttt{bind} can be applied only to existing artifacts, every grounded claim is anchored to executable evidence. This runtime check enforces provenance, not semantic truth: a misleading description of an existing artifact would still be exposed as a specific grounding edge that must be inspected by the downstream judge or auditor.

\paragraph{Derivational Expansion.}
The \texttt{infer} primitive externalizes the agent's natural-language reasoning over established claims. Rather than introducing new executable evidence, it explicitly records how a higher-level conclusion is derived from a set of premises. Given premises $\mathcal{P} \subseteq \mathcal{V}_{\mathrm{claim}}$, a reasoning annotation $r$, and a derived conclusion $c$:
\begin{equation}
\label{eq:derive_expansion}
v_{\mathrm{new}} = \texttt{infer}\bigl(\mathcal{P},\; r,\; c\bigr), \quad \Delta\mathcal{G}_t^{\mathrm{derive}} = \Bigl(\{v_{\mathrm{new}}\},\;\; \bigl\{(p, v_{\mathrm{new}}) \mid p \in \mathcal{P}\bigr\}\Bigr).
\end{equation}
This renders each derivational step explicit and auditable.

\paragraph{Terminal Extraction.}
Generation concludes when the agent invokes a dedicated \texttt{submit\_answer($\mathcal{V}_{\mathrm{final}}$)} primitive that designates a subset of established claims as terminal nodes:
\begin{equation}
\label{eq:terminal}
\mathcal{V}_{\mathrm{final}} \subseteq \mathcal{V}_{\mathrm{claim}}, \quad \mathcal{G}^{*} = \mathrm{Ancestors}_{\mathcal{G}}(\mathcal{V}_{\mathrm{final}}), \quad o = \mathrm{Compose}(\mathcal{V}_{\mathrm{final}},\; \mathcal{G}^{*}).
\end{equation}
The ancestor subgraph $\mathcal{G}^{*}$ induced by backward traversal from $\mathcal{V}_{\mathrm{final}}$ retains only the nodes and edges supporting the final response, yielding a compact evidence chain. Because \texttt{bind} references only existing variables and \texttt{infer} only prior claims, $\mathcal{G}$ remains acyclic throughout execution.

%好像缺了rollout过程
% gt -> gt*
\subsection{Graph-Based Policy Optimization}
\label{sec:training}

Given the evidence-graph construction process above, the central training question is how to make the agent learn not only to answer the task, but also to leave behind a faithful support graph. Standard outcome-supervised RL provides only a terminal signal, e.g., whether the final answer is correct. For our setting, this signal is under-specified: the intermediate objects that determine traceability are evaluated only indirectly. Therefore, a rollout may receive similar final-answer feedback despite failed executions, unsupported semantic jumps, or incomplete provenance chains, making it difficult to assign credit to graph-building decisions that actually determine evidence quality.

We address this with graph-based policy optimization, which aligns reward assignment with the order in which the graph is constructed, as illustrated in Figure~\ref{fig:main_fig}(c). For each rollout that produces $\mathcal{G}_T$, training observes three objects created by the policy: the action sequence, the derivational expansions, and the extracted terminal evidence subgraph. We attach rewards to these objects at three corresponding granularities: action-level execution rewards supervise the computational backbone, edge-level verification rewards supervise \texttt{infer} expansions, and a terminal outcome reward supervises whether the selected evidence subgraph supports the final answer. 
%The resulting training--inference asymmetry is intentional: auxiliary verifiers provide fine-grained supervision during RL, while inference uses the learned policy without these reward models.

\paragraph{Cold-Start via Trajectory Distillation.}
Base models are pretrained with result-oriented objectives and cannot natively emit the intermediate evidence graphs our framework requires, leaving the RL reward signal too sparse to optimize from such an initialization. We therefore introduce a cold-start stage that distills trajectories from a strong teacher operating inside the full \ours runtime, yielding roughly 36K supervised examples after rejection sampling and rule-based filtering.

We organize these examples at two complementary granularities. The first granularity consists of \emph{atomic samples}, each isolating a single primitive that the agent must master, including next-action prediction under compressed observations, first-step planning from the user query, and final report generation from a completed evidence subgraph. The second granularity consists of \emph{full trajectories}, which preserve the end-to-end construction of an evidence graph and expose the model to long-horizon dependencies. To keep the training distribution faithful to inference, each retained trajectory is replayed through the runtime so that its intermediate results are recovered under the same sliding-window compression used at deployment.

Training proceeds as a two-stage curriculum. The atomic stage shuffles the three types of samples so that the model internalizes the syntax and post-conditions of each primitive without entanglement. The trajectory stage then composes these primitives into coherent multi-turn graph construction. This separation also enables the ablation in \S\ref{sec:ablation} to attribute gains to the two stages independently. Further details on data construction and training are provided in Appendices~\ref{app:sft-data} and~\ref{app:training}.

\begin{table*}[t]
\centering
\caption{Performance on four data-intensive benchmarks. \textbf{Traceability} reflects an output's structural capacity for evidence tracing: \textbf{Comp.}\ (computational provenance) and \textbf{Deriv.}\ (derivational provenance). \textbf{GR} measures claim support recoverable from the method's exposed evidence artifact.}
\label{tab:main_acc}
\footnotesize
\setlength{\tabcolsep}{2.5pt}
\renewcommand{\arraystretch}{1.15}
\begin{tabular}{@{}l cc cc cccc cc c c@{}}
\toprule
\multirow{2}{*}{\textbf{Method}} & \multicolumn{2}{c}{\textbf{Traceability}} & \multirow{2}{*}{DABench} & \multirow{2}{*}{DSBench} & \multicolumn{4}{c}{TableBench} & \multicolumn{2}{c}{DABStep-R} & {\multirow{2}{*}{\textbf{Overall}}} & {\multirow{2}{*}{\textbf{GR}}} \\
\cmidrule(lr){2-3}\cmidrule(lr){6-9}\cmidrule(lr){10-11}
 & Comp. & Deriv. & & & Fact & Num. & Anal. & Avg. & Content & Format & & \\
\midrule
\rowcolor{TableGroup}
\multicolumn{13}{@{}l}{\textit{\textbf{Direct inference}}} \\
GPT-5.2        & \tracenone & \tracenone & 19.46 & 14.72 & 85.42 & 77.83 & 48.40 & 66.63 & 4.61 & 4.81 & 48.75 & \pending \\
Gemini-2.5-Pro & \tracenone & \tracenone & 31.91 & 27.61 & 82.29 & 82.62 & 45.19 & 67.22 & 4.82 & 4.85 & 55.86 & \pending \\
Claude-4.5-Sonnet & \tracenone & \tracenone & 32.30 & 25.77 & 87.50 & 85.39 & 53.64 & 72.61 & 4.80 & 4.64 & 56.27 & \pending \\
Claude-4.5-Opus   & \tracenone & \tracenone & 36.58 & 30.67 & 84.38 & 87.41 & 55.69 & 74.04 & 4.68 & 4.73 & 58.85 & \pending \\
Qwen3-32B    & \tracenone & \tracenone & 22.57 & 29.75 & 88.54 & 84.13 & 48.40 & 69.98 & 4.65 & 4.64 & 53.80 & \pending \\
Qwen3-30B-A3B    & \tracenone & \tracenone & 26.46 & 26.69 & 84.38 & 85.89 & 46.06 & 69.38 & 4.49 & 4.53 & 53.18 & \pending \\
\midrule
\rowcolor{TableGroup}
\multicolumn{13}{@{}l}{\textit{\textbf{ReAct data agents}}} \\
GPT-5.2           & \tracefull & \tracenone & 80.93 & 44.48 & 75.00 & 79.09 & 52.48 & 67.70 & 3.26 & 3.42 & 64.98 & 74.17 \\
GPT-5.4           & \tracefull & \tracenone & 87.16 & 53.07 & 77.08 & 84.63 & 62.10 & 74.52 & 3.71 & 3.62 & 72.01 & \second{78.52} \\
Claude-4.5-Sonnet & \tracefull & \tracenone & 86.77 & 44.79 & 72.92 & 72.07 & 43.01 & 62.52 & 4.55 & 4.51 & 71.17 & 74.16 \\
Claude-4.5-Opus   & \tracefull & \tracenone & 89.49 & 53.07 & 86.15 & 86.46 & 57.43 & 74.40 & 3.67 & 3.92 & \second{73.22} & 73.57 \\
Qwen3-8B          & \tracefull & \tracenone & 80.54 & 38.04 & 60.42 & 72.54 & 52.19 & 62.80 & 3.39 & 3.41 & 62.35 & 59.52 \\
Qwen3-Coder-30B & \tracefull & \tracenone & 58.37 & 28.53 & 65.62 & 60.15 & 35.86 & 50.78 & 3.21 & 3.52 & 51.25 & 73.86 \\
QwQ-32B           & \tracefull & \tracenone & 80.54 & 46.01 & 77.08 & 70.03 & 51.31 & 63.16 & 3.20 & 3.20 & 63.43 & 69.89 \\
Qwen3-32B       & \tracefull & \tracenone & 84.82 & 56.44 & 78.12 & 79.09 & 50.73 & 67.34 & 3.81 & 3.72 & 70.98 & 71.95 \\
Qwen3-30B-A3B   & \tracefull & \tracenone & 83.66 & 47.24 & 68.75 & 72.54 & 51.60 & 63.52 & 3.09 & 3.21 & 64.36 & 64.08 \\
\midrule
\rowcolor{TableGroup}
\multicolumn{13}{@{}l}{\textit{\textbf{Specialized data agents}}} \\
DataMind     & \tracefull & \tracenone & 83.27 & 14.72 & 73.96 & 82.12 & 50.73 & 68.94 & 2.47 & 2.86 & 55.06 & 71.64 \\
DeepAnalyze  & \tracefull & \tracenone & 75.60 & 47.64 & 80.21 & 71.92 & 50.40 & 67.51 & 3.61 & 3.93 & 66.54 & 61.35 \\
\rowcolor{OursRow}
\textbf{\ours-8B} & \tracefull & \tracefull & 85.99 & 66.43 & 87.23 & 83.63 & 57.78 & 73.58 & 3.31 & 3.56 & \best{73.68} & \best{87.61} \\
\bottomrule
\end{tabular}
% \vspace{-0.4em}
\end{table*}

\paragraph{Graph-Aware Reward Design.}
Concretely, the three rewards supervise progressively coarser objects produced by the same rollout: actions $\{\alpha_t\}_{t=1}^T$, derivational expansions $\mathcal{I} = \{(\mathcal{P}_i, r_i, c_i)\}_i$, and the terminal evidence subgraph $\mathcal{G}^{*}$ from Eq.~\ref{eq:terminal}. We therefore use a composite reward that follows the construction order of the graph itself:
\begin{equation}
\label{eq:reward}
R(\tau, \mathcal{G}_T) = \underbrace{R_{\text{process}}(\{\alpha_t\}_{t=1}^T)}_{\text{computational integrity}} + \underbrace{R_{\text{infer}}(\mathcal{I})}_{\text{derivational validity}} + \underbrace{R_{\text{outcome}}(\mathcal{G}^{*}, q, a^*)}_{\text{terminal subgraph quality}}.
\end{equation}
The relation among the three terms is hierarchical: $R_{\text{process}}$ keeps the computational backbone executable, which is a prerequisite for valid \texttt{bind} operations; $R_{\text{infer}}$ then checks whether the resulting claim graph contains justified semantic transitions; $R_{\text{outcome}}$ finally scores whether the selected terminal subgraph actually answers the task faithfully. We do not introduce a separate grounding reward: a \texttt{bind} call is accepted only when it references an existing runtime artifact, so grounding validity is enforced structurally by the environment.

The \emph{process} term operates at action level and aggregates per-step execution feedback over the computational backbone:
\begin{equation}
\label{eq:process_reward}
R_{\text{process}} = \frac{1}{T}\sum_{t=1}^{T} \mathbb{I}[\texttt{exec}(\alpha_t) = \text{success}].
\end{equation}
This term encourages the policy to produce executable code paths and stable intermediate variables, rather than reaching the correct answer through brittle or partially failed traces.

The \emph{inference} term operates at the derivational-edge level. For each \texttt{infer} invocation $i = (\mathcal{P}_i, r_i, c_i) \in \mathcal{I}$, an external verifier checks whether conclusion $c_i$ follows from premises $\mathcal{P}_i$ under reasoning annotation $r_i$:
\begin{equation}
\label{eq:infer_reward}
R_{\text{infer}} = \frac{1}{|\mathcal{I}|}\sum_{i \in \mathcal{I}} \texttt{Verify}(q, \mathcal{P}_i, r_i, c_i), \quad \texttt{Verify}(\cdot) \in \{-0.5, +1\}, \quad R_{\text{infer}} := 0 \;\;\text{if}\;\; \mathcal{I} = \emptyset.
\end{equation}
This term penalizes unsupported semantic jumps even when the final answer happens to be correct.

The \emph{outcome} term operates at the graph-answer level. An LLM-as-judge scores the extracted terminal subgraph together with the final answer against a task-specific rubric (Appendix~\ref{app:eval}):
\begin{equation}
\label{eq:outcome_reward}
R_{\text{outcome}} = \mathbb{I}[\text{terminal extraction}] \cdot \texttt{Judge}(q,\, \mathcal{G}^{*},\, a^*) \,/\, S,
\end{equation}
where $\texttt{Judge}(\cdot)$ evaluates answer correctness, completeness, and faithfulness, and $S$ is the maximum rubric score.

We optimize the policy using DAPO~\cite{dapo}, treating $R(\tau, \mathcal{G}_T)$ as the trajectory-level return:
\begin{equation}
\label{eq:objective}
\mathcal{J}(\theta) = \mathbb{E}_{(q, \mathcal{F}, d) \sim \mathcal{D}} \left[\frac{1}{N}\sum_{n=1}^{N} \sum_{t} \min\!\left(\frac{\pi_\theta}{\pi_{\theta_{\text{old}}}} \hat{A}^{(n)},\; \text{clip}\!\left(\frac{\pi_\theta}{\pi_{\theta_{\text{old}}}}, 1{-}\epsilon_l, 1{+}\epsilon_h\right)\hat{A}^{(n)}\right)\right],
\end{equation}
where $N$ is the group size and $\hat{A}^{(n)}$ the group-normalized advantage from $R(\tau^{(n)}, \mathcal{G}_T^{(n)})$; per-tuple verifier overhead is $\mathcal{O}(N(|\mathcal{I}|+1))$, kept within $\sim 1.3{\times}$ outcome-only rollout cost via batching and a small dedicated verifier (Appendix~\ref{app:training}).

\section{Experiments}

\subsection{Experimental Setup}
\label{sec:exp-setup}

\paragraph{Benchmarks and Metrics.}
We evaluate \ours on four data-intensive benchmarks covering three task types and multiple domains. \textit{Table QA}: \textbf{TableBench}~\cite{tablellm} ($\sim$700 single-table questions over fact-checking, numerical reasoning, and data-analysis subsets). \textit{Data Analysis}: \textbf{InfiAgent-DABench}~\cite{infiagentdabench} (257 single-CSV questions) and \textbf{DSBench}~\cite{dsbench} (466 multi-table tasks with long contexts). \textit{Multi-step Research}: \textbf{DAB-Step Research}, a 100-case subset we curate from DABstep~\cite{dabstep}, in which each task jointly reasons over tables and unstructured documentation. Accuracy on QA and data-analysis tasks is judged by an LLM. For research tasks, an LLM judge scores each output on \textit{Content} and \textit{Format}, following prior work~\cite{deepanalyze}. We treat evaluation as two-dimensional: benchmark scores measure answer correctness and completeness, while \textbf{Grounding Rate (GR)} measures whether the answer's stated claims are recoverable from the evidence artifact exposed by each method.  Detailed protocols and prompts are in Appendix~\ref{app:traceability-eval}, with a cross-model consistency check for the LLM-based judgments in Appendix~\ref{app:cross-model-consistency}.

\paragraph{Baselines.} We compare against three families of methods: (1) \textit{Direct inference}: LLMs are fed the input files and produce final answers; (2) \textit{ReAct data agents}~\cite{react,wang2024codeact}: equip the agent with a Python tool in a ReAct loop; (3) \textit{Specialized data agents}: data agents trained for data-intensive tasks, including DataMind~\cite{datamind} and DeepAnalyze~\cite{deepanalyze}. See Appendix~\ref{app:baselines} for details.

\paragraph{Implementation.} Main results use Qwen3-8B~\cite{qwen3} as backbone, with at most 50 interaction turns, $8{,}192$ generation tokens, and a $32{,}768$-token context. SFT runs on $36$K instances with MS-Swift~\cite{msswift} (max sequence length $32{,}768$, learning rate $1\mathrm{e}{-5}$). RL uses Verl~\cite{verl} with DAPO~\cite{dapo}, $8$ rollouts per prompt and learning rate $1\mathrm{e}{-6}$. Full details are in Appendix~\ref{app:training}.

\subsection{Main Results}
\begin{figure}[!t]
\centering
% Subfigure (a): combined ablation + robustness table
\begin{subfigure}[t]{0.58\textwidth}
\vspace{0pt}
\centering
\setlength{\tabcolsep}{2.0pt}
\fontsize{7.2pt}{9.1pt}\selectfont
\begin{tabular}{@{}l ccccc cc@{}}
\toprule
\textbf{Variant} & Infi. & DS & Table & DAB & \textbf{Overall} & \textbf{GR} & $\Delta$ \\
\midrule
\rowcolor{OursRow}
\textbf{Full \ours-8B} & 85.99 & 66.43 & 73.58 & 3.31 & 73.68 & 87.61 & --- \\
\midrule
\rowcolor{TableGroup}
\multicolumn{8}{@{}l}{\textit{\textbf{Training recipe}}} \\
\quad w/o atomic SFT      & 82.75 & 57.28 & 67.61 & 3.49 & 69.34 & 72.33 & $-4.34$ \\
\quad w/o traj.\ SFT       & 26.46 & 69.63 & \phantom{0}6.22 & 2.44 & 37.78 & 95.01 & $-35.90$ \\
\quad w/o RL stage        & 85.10 & 62.40 & 71.85 & 3.12 & 70.42 & 85.29 & $-3.26$ \\
\quad outcome-only RL        & 84.05 & 65.52 & 64.62 & 2.15 & 65.35 & 76.46 & $-8.33$ \\
\midrule
\rowcolor{TableGroup}
\multicolumn{8}{@{}l}{\textit{\textbf{Backbone size}}} \\
\quad \ours-4B            & 81.25 & 65.91 & 68.44 & 2.88 & 68.28 & 69.41 & $-5.40$ \\
\quad \ours-14B           & 88.98 & 70.16 & 75.74 & 3.28 & 75.52 & 88.67 & $+1.84$ \\
\bottomrule
\end{tabular}
\end{subfigure}
\hfill
% Subfigure (b): traceability figure
\begin{subfigure}[t]{0.4\textwidth}
\vspace{7pt}
\centering
\includegraphics[width=\linewidth]{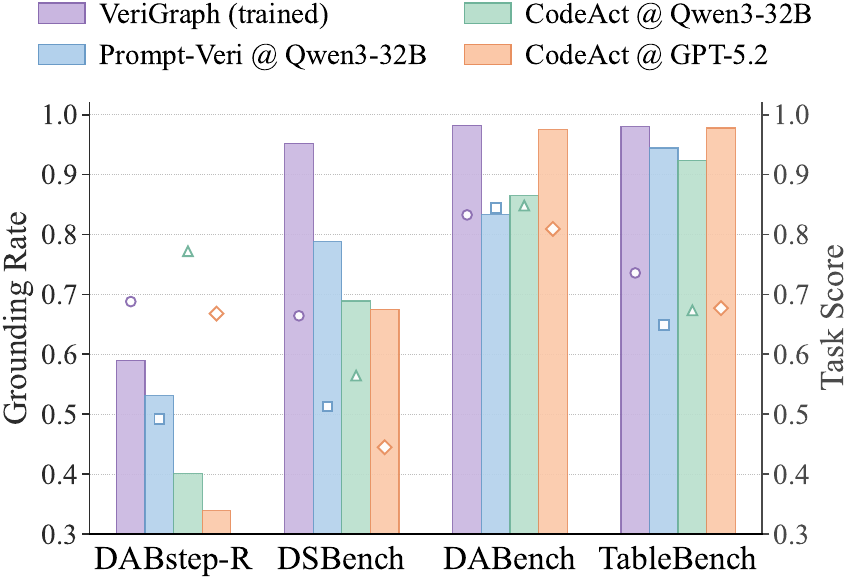}
\end{subfigure}
\caption{\textbf{Left}: ablation on various components and robustness across backbone sizes. \textbf{Right}: Grounding Rate vs.\ task performance across four traceability settings, with task performance shown as points and Grounding Rate shown as bars.}
\label{fig:ablation_trace}
\vspace{-0.8em}
\end{figure}

% Table~\ref{tab:main_acc} compares \ours with direct inference, ReAct data agents, and specialized data agents across four benchmarks.
% \textbf{(1) Evidence-graph construction improves the accuracy--verifiability frontier.} Built on Qwen3-8B, \ours achieves the best Overall score ($73.68$), outperforming its same-backbone ReAct counterpart by $+11.33$ points and slightly surpassing the strongest proprietary ReAct baseline, Claude-4.5-Opus, while providing both computational and derivational provenance.
% \textbf{(2) The gains concentrate on long-evidence tasks.}
% Compared with Qwen3-30B-A3B ReAct, \ours improves DSBench by $+19.19$ and TableBench by $+10.06$, suggesting that explicit graph state helps preserve, reuse, and reconcile intermediate computations across multi-step analysis.
% \textbf{(3) Higher task performance is accompanied by stronger claim support.}
% \ours obtains the highest GR ($87.61$), showing that its final claims are more consistently supported by upstream computations and derivations than claims extracted from linear ReAct transcripts.
\textbf{(1) Reaching the proprietary frontier under a stronger evidence contract.} With only $8$B parameters, \ours matches Claude-4.5-Opus ReAct on Overall ($73.68$ vs.\ $73.22$). Crucially, this parity is reached under a strictly stronger evidence requirement: \ours is the only entry in Table~\ref{tab:main_acc} that simultaneously exposes computational and derivational provenance, and it attains the highest GR ($87.61$, $+14.04$ over the strongest ReAct baseline).
\textbf{(2) Evidence quality holds even where surface-level scoring is harshest.}
On DAB-Step Research, \ours attains the highest GR among all systems, despite Content/Format scores ($3.31/3.56$) that trail proprietary direct-inference baselines ($\geq\!4.6$). The gap stems from the answer being serialized from the terminal evidence subgraph rather than written as free-form prose, which the LLM judge tends to favor in presentation. The same effect is visible across all ReAct-style agents (e.g., Claude-4.5-Opus drops from $4.68/4.73$ under direct inference to $3.67/3.92$ under ReAct), indicating a stylistic penalty on structured outputs rather than a deficit in evidence support.

\subsection{Traceability Analysis}
\label{sec:traceability}

Figure~\ref{fig:ablation_trace} (right) isolates the relation between task performance and claim grounding. \textbf{(1) Reliable grounding is not a by-product of tool use.} CodeAct agents can reach competitive task scores, yet their GR remains substantially below \ours, showing that a correct-looking trajectory may still leave many final claims hard to audit. \textbf{(2) Prompted structure helps but is insufficient.} Prompt-Veri improves over flat CodeAct on grounding, but still trails the trained \ours policy, suggesting that models must learn when and how to materialize evidence rather than only be instructed to output graph-like traces. \textbf{(3) \ours improves the accuracy--traceability tradeoff.} It pairs the strongest aggregate GR with strong task performance, supporting explicit evidence-graph construction as a practical route to verifiable data analysis.

% --- 6.3 Ablation Studies (~1/2 page) ---
\subsection{Ablation Studies}
\label{sec:ablation}

Figure~\ref{fig:ablation_trace} (left) ablates the training recipe behind evidence-graph construction. \textbf{(1) Trajectory SFT is the enabling stage.} Removing it reduces Overall by $35.90$ points and collapses TableBench from $73.58$ to $6.22$. This indicates that primitive-level imitation can make the model follow the graph interface well, as reflected by its high GR ($95.01$), but it does not teach the long-horizon composition of code execution, \texttt{bind}, and \texttt{infer} needed to solve tasks. The result illustrates why GR is evaluated jointly with task performance: locally grounded claims do not imply a complete or useful answer. \textbf{(2) Atomic SFT stabilizes the graph interface.} Omitting it yields smaller but consistent drops in Overall ($-4.34$) and GR (to $72.33$), suggesting that short primitive-level examples teach the local preconditions and post-conditions needed for well-typed graph operations. \textbf{(3) Graph-aware RL is needed for verifiable gains.} The SFT-only policy remains strong ($70.42$ Overall, $85.29$ GR), while outcome-only RL degrades both accuracy and grounding ($65.35$ Overall, $76.46$ GR). This gap shows that final-answer rewards can misalign optimization with evidence quality. Our composite reward instead assigns credit to computation, evidence selection, and derivational validity.

% ============================================================================
% 6. ANALYSIS & DISCUSSION (~0.75 page)
% ============================================================================
\section{Analysis}
\subsection{Graph Interpretability}

\begin{figure}[!t]
\centering
\includegraphics[width=\linewidth]{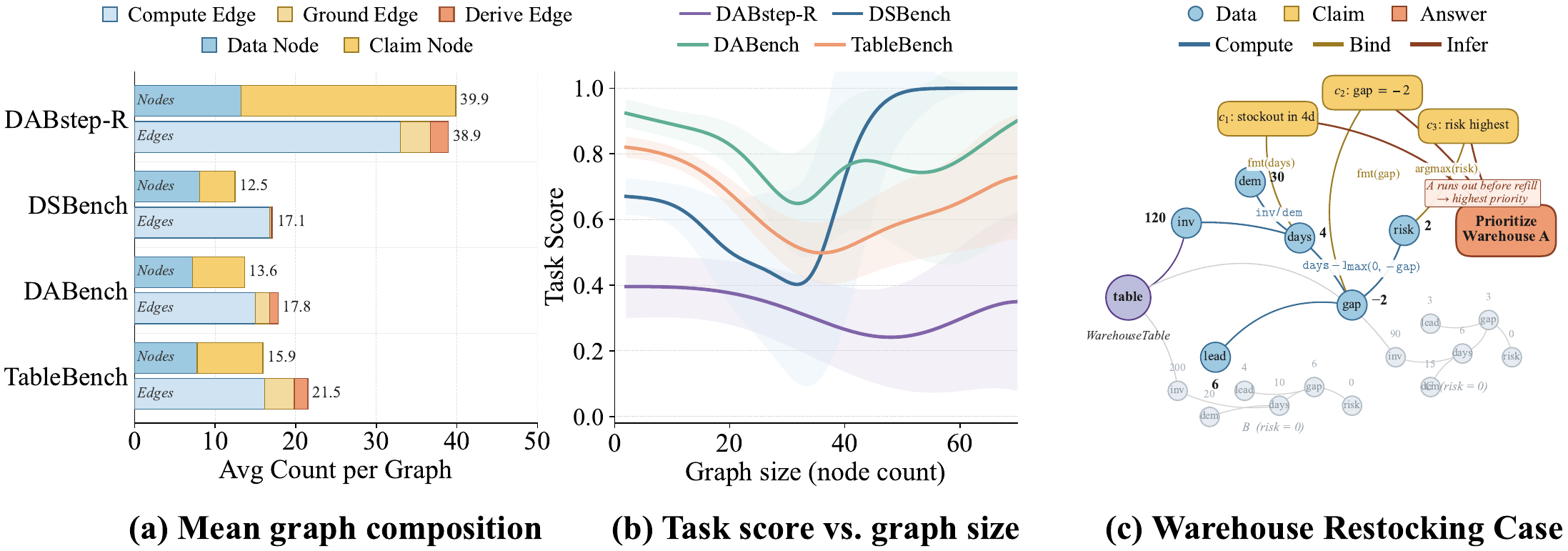}
\caption{Evidence-graph analysis. \textbf{(a)} Mean node and edge composition per benchmark. \textbf{(b)} Smoothed task score versus graph size with uncertainty bands. \textbf{(c)} Example warehouse-restocking DAG with the selected answer chain highlighted.}
\label{fig:graph_analysis}
\vspace{-0.5em}
\end{figure}

Figure~\ref{fig:graph_analysis} analyzes the terminal evidence graphs produced by \ours. \textbf{(1) Graph topology reflects the task regime.} DAB-Step Research yields the largest graphs because many computed signals must be promoted into narrative claims. By contrast, data-analysis benchmarks are dominated by computational edges, while TableBench has a denser claim layer for semantic comparison over tabular facts. \textbf{(2) Graph size provides an internal difficulty signal.} Task scores decline as terminal graphs grow, indicating that larger graphs mainly reflect evidence burden, including additional quantities, joins, comparisons, and cross-source reconciliation, rather than superficial verbosity. \textbf{(3) Auditability becomes local.} In the warehouse case, the final recommendation can be traced backward through typed computation, grounding, and derivation edges. This structure localizes potential failures to a specific calculation, grounded statement, or inference step, instead of forcing reviewers to inspect a long linear transcript.

\subsection{Robustness Across Backbones}
\label{sec:robustness}

Applying the same recipe to Qwen3-4B and Qwen3-14B (Figure~\ref{fig:ablation_trace}, left) shows that the benefit is not tied to a single backbone size. \textbf{(1) The graph interface transfers across scales.} Even \ours-4B surpasses the vanilla Qwen3-8B ReAct agent in Overall score, indicating that the gains cannot be explained by capacity alone. \textbf{(2) Scaling helps but quickly saturates.} Moving from 4B to 8B yields a larger gain ($+5.40$) than moving from 8B to 14B ($+1.84$), placing the 8B model near a favorable accuracy--cost frontier. \textbf{(3) Different tasks stress different bottlenecks.} TableBench continues to improve with model size, whereas DAB-Step Research changes little from 8B to 14B, suggesting that report-style synthesis depends on evidence coverage in addition to raw model capacity.

% ============================================================================
% 7. CONCLUSION (~0.25 page)
% ============================================================================
\section{Conclusion}

We presented \ours, a neuro-symbolic framework that recasts data-intensive agent reasoning as the incremental construction of a heterogeneous evidence DAG. Through computational, grounding, and derivational expansion primitives embedded in the code action space, \ours unifies deterministic computation with semantic deduction and reduces structural traceability to graph reachability. Built on Qwen3-8B, \ours achieves the highest Overall score among evaluated baselines while producing auditable evidence graphs. These results suggest that evidence structure should be treated as a first-class optimization target, not merely as a post-hoc explanation attached to an answer.

\bibliographystyle{plainnat}
\bibliography{reference}

% ============================================================================
% APPENDIX (does not count toward page limit)
% ============================================================================
% \newpage
% \input{checklist.tex}

\appendix

\newpage
\section*{Appendix}
\startcontents[sections]
\setupappendixtoc
\printcontents[sections]{l}{1}{\setcounter{tocdepth}{3}}

\section{VeriGraph Runtime Details}
\label{app:verigraph}

This section complements the formal exposition in Section~\ref{sec:method} with a
self-contained description of the \ours runtime that is shared by inference, SFT
data synthesis, and RL training. Three design points jointly determine the
behaviour of the system and are the focus of this appendix:
(i)~the rollout reinterpreted as graph expansion (Appendix~\ref{app:verigraph-algorithm}),
(ii)~the formal semantics of the \texttt{bind}, \texttt{infer}, and
\texttt{submit\_answer} primitives that the policy invokes inside its code action
(Appendix~\ref{app:verigraph-graph}),
and (iii)~the structured per-turn observation that summarises the executor and
graph state under a bounded context budget
(Appendix~\ref{app:verigraph-context}).
We close with the integration of this runtime into the RL training loop
(Appendix~\ref{app:verigraph-rl-impl}).

\subsection{Graph-Augmented Rollout}
\label{app:verigraph-algorithm}

We adopt the notation of Section~\ref{sec:method} without change: $q$ is the
user query, $\mathcal{F}$ the attached data sources, $\pi_\theta$ the policy,
$\mathcal{E}$ the Python interpreter, $\mathcal{V}_t$ the executor namespace,
$\mathcal{G}_t = (\mathcal{V}_{\mathrm{data}}\cup\mathcal{V}_{\mathrm{claim}},\,\mathcal{E}_{\mathrm{comp}}\cup\mathcal{E}_{\mathrm{ground}}\cup\mathcal{E}_{\mathrm{derive}})$
the heterogeneous evidence DAG of Eq.~\eqref{eq:graph}, and
$S_t = (H_t,\mathcal{V}_t,\mathcal{G}_{t-1})$ the structured observation.
Algorithm~\ref{alg:verigraph} formalises a single rollout. It differs from a
vanilla CodeAct loop in three respects: (i)~the executor is initialised with a
graph runtime exposing the three primitives of Eqs.~\eqref{eq:ground_expansion}--\eqref{eq:terminal};
(ii)~each tool response is rendered together with a structured summary of
$\mathcal{V}_t$ and $\mathcal{G}_t$ rather than only \texttt{stdout}; and
(iii)~termination is controlled by the graph state via
\texttt{submit\_answer} rather than by a textual \texttt{<answer>} tag.

\begin{algorithm}[htbp]
\caption{\ours rollout. The policy interacts with a Python interpreter $\mathcal{E}$
endowed with the graph primitives $\Pi=\{\texttt{bind},\texttt{infer},\texttt{submit\_answer}\}$
and returns the terminal evidence subgraph $\mathcal{G}^{*}$ together with the
set of submitted final claims $\mathcal{V}_{\mathrm{final}}\subseteq\mathcal{V}_{\mathrm{claim}}$.}
\label{alg:verigraph}
\begin{algorithmic}[1]
\REQUIRE Query $q$; files $\mathcal{F}$; policy $\pi_\theta$; horizon $T$; recent-window size $k$; per-field budget $b$
\ENSURE Terminal subgraph $\mathcal{G}^{*}$ and final claim set $\mathcal{V}_{\mathrm{final}}$
\STATE $\mathcal{G}_0 \leftarrow (\emptyset,\emptyset)$, $\mathcal{V}_0 \leftarrow \mathcal{E}.\textsc{Init}(\mathcal{F},\Pi)$
\STATE $H_0 \leftarrow (\textsc{SysPrompt},\,q\oplus\mathcal{F})$
\FOR{$t=1,\dots,T$}
    \STATE $S_t \leftarrow (H_{t-1},\,\mathcal{V}_{t-1},\,\mathcal{G}_{t-1})$ \hfill\COMMENT{observation; cf.\ Sec.~\ref{sec:primitives}}
    \STATE $(\tau_t,\alpha_t) \sim \pi_\theta(\cdot \mid S_t)$ \hfill\COMMENT{thought \& code action}
    \IF{$\alpha_t = \emptyset$} \STATE \textbf{break} \hfill\COMMENT{degenerate; no graph update} \ENDIF
    \STATE $z_t,\,\Delta\mathcal{V}_t,\,\Delta\mathcal{G}_t \leftarrow \mathcal{E}.\textsc{Exec}(\alpha_t)$ \hfill\COMMENT{$\Delta\mathcal{G}_t=\Delta\mathcal{G}_t^{\mathrm{comp}}\cup\Delta\mathcal{G}_t^{\mathrm{ground}}\cup\Delta\mathcal{G}_t^{\mathrm{derive}}$, Eq.~\eqref{eq:graph_update}}
    \STATE $\mathcal{V}_t \leftarrow \mathcal{V}_{t-1}\cup\Delta\mathcal{V}_t$,\quad $\mathcal{G}_t \leftarrow \mathcal{G}_{t-1}\cup\Delta\mathcal{G}_t$
    \STATE $\tilde z_t \leftarrow \textsc{Render}(z_t,\,\mathcal{V}_t,\,\mathcal{G}_t;\,b)$ \hfill\COMMENT{tool response, per-field budget $b$}
    \STATE $H_t \leftarrow \textsc{Compress}\bigl(H_{t-1}\oplus(\alpha_t,\tilde z_t);\,k\bigr)$ \hfill\COMMENT{recent-$k$, Eq.~\eqref{eq:compress}}
    \IF{\textsc{Submitted}$(\mathcal{G}_t)$} \STATE \textbf{break} \ENDIF
\ENDFOR
\STATE $\mathcal{V}_{\mathrm{final}} \leftarrow \{v\in\mathcal{V}_{\mathrm{claim}} : v.\mathrm{final}=1\}$,\quad $\mathcal{G}^{*}\leftarrow\mathrm{Ancestors}_{\mathcal{G}_t}(\mathcal{V}_{\mathrm{final}})$
\RETURN $\mathcal{G}^{*}$,\;$\mathcal{V}_{\mathrm{final}}$
\end{algorithmic}
\end{algorithm}

The terminal output $o$ is obtained from $\mathcal{V}_{\mathrm{final}}$ and
$\mathcal{G}^{*}$ via the $\mathrm{Compose}$ map of Eq.~\eqref{eq:terminal}.
For report tasks ($|\mathcal{V}_{\mathrm{final}}|>1$ at inference), $\mathrm{Compose}$
is instantiated as a constrained writer that reorders the final claims into
prose without introducing new graph nodes, so $\mathcal{G}^{*}$ remains the
sole carrier of evidence. RL rollouts skip this post-hoc writer and optimise
directly against $\mathcal{V}_{\mathrm{final}}$.

\subsection{Runtime Realisation of the Graph Primitives}
\label{app:verigraph-graph}

The three primitives \texttt{bind}, \texttt{infer}, and \texttt{submit\_answer}
are defined in Eqs.~\eqref{eq:ground_expansion}--\eqref{eq:terminal}. Here we
specify how the runtime exposes them to the policy and what local checks it
applies to keep $\mathcal{G}$ well-formed. We retain all symbols from
Section~\ref{sec:method}: $\mathcal{V}_{\mathrm{data}}$ for data nodes,
$\mathcal{V}_{\mathrm{claim}}$ for claim nodes, $\mathcal{V}_{\mathrm{final}}$
for the submitted final claims, and $\mathcal{G}^{*}$ for the terminal
evidence subgraph.

\paragraph{\texttt{bind} (Eq.~\eqref{eq:ground_expansion}).}
The implementation generalises the single-source signature
$v_c=\texttt{bind}(v_d,l)$ to a multi-source one. The agent supplies a templated
sentence $l$ containing placeholders $\{x_j\}_{j=1}^{m}$ together with a binding
$\{x_j \mapsto v_d^{(j)}\}$ that assigns each placeholder to an existing
data node $v_d^{(j)}\in\mathcal{V}_{\mathrm{data}}$. The runtime renders
$\mathrm{content}(v_c)=l[\,x_j\mapsto\mathrm{val}(v_d^{(j)})\,]$ and creates the
grounding update
\begin{equation}
  v_c \in \mathcal{V}_{\mathrm{claim}}, \qquad
  \Delta\mathcal{G}_t^{\mathrm{ground}} = \Bigl(\{v_c\},\;\bigl\{(v_d^{(j)},\,v_c)\bigr\}_{j=1}^{m}\Bigr),
\end{equation}
which reduces to Eq.~\eqref{eq:ground_expansion} when $m=1$. A local check
rejects templates with no placeholders, so each atomic claim must expose an
explicit link to executor state. This check enforces referential anchoring, but
does not by itself prove that the surrounding natural-language description is a
faithful characterization of the bound values.

\paragraph{\texttt{infer} (Eq.~\eqref{eq:derive_expansion}).}
The operator $v_{\mathrm{new}}=\texttt{infer}(\mathcal{P},r,c)$ takes premises
$\mathcal{P}\subseteq\mathcal{V}_{\mathrm{claim}}$, a reasoning string $r$, and
a conclusion $c$, and produces a derived claim
$v_{\mathrm{new}}\in\mathcal{V}_{\mathrm{claim}}$ with
$\mathrm{content}(v_{\mathrm{new}})=c$ and $\mathrm{reasoning}(v_{\mathrm{new}})=r$,
inserting the edges of Eq.~\eqref{eq:derive_expansion} into
$\mathcal{E}_{\mathrm{derive}}$. The runtime rejects non-claim premises, empty
$c$ or $r$, and reasoning strings exceeding a fixed budget, so each
$\mathcal{E}_{\mathrm{derive}}$ edge is anchored to existing graph content
rather than to free executor state.

\paragraph{\texttt{submit\_answer} (Eq.~\eqref{eq:terminal}).}
$\texttt{submit\_answer}(\mathcal{V}_{\mathrm{final}})$ flags each
$v\in\mathcal{V}_{\mathrm{final}}\subseteq\mathcal{V}_{\mathrm{claim}}$ as
terminal and signals the rollout loop to halt after the current code cell
returns. The exporter then materialises the terminal evidence subgraph
$\mathcal{G}^{*}=\mathrm{Ancestors}_{\mathcal{G}}(\mathcal{V}_{\mathrm{final}})$
defined in Eq.~\eqref{eq:terminal}; subsequent code cells, if any, cannot
mutate $\mathcal{G}$.

\paragraph{Provenance and traceability.}
Algorithm~\ref{alg:verigraph} updates $\mathcal{G}$ monotonically, since every
edge enters $\mathcal{G}_t$ either through the automatic computational
expansion $\Delta\mathcal{G}_t^{\mathrm{comp}}$ of Eq.~\eqref{eq:comp_expansion}
or through the agent-invoked expansions of Eqs.~\eqref{eq:ground_expansion}
and~\eqref{eq:derive_expansion}. Together with the type constraints
$\mathcal{E}_{\mathrm{comp}}\subseteq\mathcal{V}_{\mathrm{data}}\times\mathcal{V}_{\mathrm{data}}$,
$\mathcal{E}_{\mathrm{ground}}\subseteq\mathcal{V}_{\mathrm{data}}\times\mathcal{V}_{\mathrm{claim}}$,
$\mathcal{E}_{\mathrm{derive}}\subseteq\mathcal{V}_{\mathrm{claim}}\times\mathcal{V}_{\mathrm{claim}}$
of Eq.~\eqref{eq:graph}, this guarantees that $\mathcal{G}^{*}$ is acyclic and
that every $v\in\mathcal{V}_{\mathrm{final}}$ admits a backward traversal in
$\mathcal{G}^{*}$ terminating at raw-data nodes in $\mathcal{V}_{\mathrm{data}}$.
This is the structural property used to materialize the evidence context for the
Grounding Rate metric (Appendix~\ref{app:traceability-eval}) and the premise
sets for the inference reward in Appendix~\ref{app:verigraph-rl-impl}. Semantic
support is then judged separately from reachability.

\paragraph{Graph storage and answer extraction.}
The runtime stores the claim layer
$\bigl(\mathcal{V}_{\mathrm{claim}},\,\mathcal{E}_{\mathrm{ground}}\cup\mathcal{E}_{\mathrm{derive}}\bigr)$
explicitly as a set of \texttt{Claim} records. Each record holds its
content, type, and premise ids. For atomic claims it additionally records
the bound executor variables $\{x_j\mapsto v_d^{(j)}\}$ and a snapshot of
their values at \texttt{bind} time. The data layer
$(\mathcal{V}_{\mathrm{data}},\mathcal{E}_{\mathrm{comp}})$ is not
mirrored into a separate store. Instead, the runtime keeps the raw inputs
loaded at $t=0$ together with the ordered code cells
$(\alpha_1^{\mathrm{code}},\dots,\alpha_T^{\mathrm{code}})$ as part of the
trajectory. Because the bound identifiers $v_d^{(j)}$ in each atomic claim
name specific variables produced by these cells, any data node referenced
from the claim layer can be located, inspected, and recomputed by
re-executing the prefix that defined it. This is enough for the structural
guarantees stated above. Every $v\in\mathcal{V}_{\mathrm{final}}$ reaches
data nodes through stored
$\mathcal{E}_{\mathrm{ground}}\cup\mathcal{E}_{\mathrm{derive}}$ edges, and
each data node is anchored to a specific cell in the executable log.
Maintaining a duplicate $\mathcal{E}_{\mathrm{comp}}$ store would have to
mirror Python semantics such as pandas mutations, in-place updates, and
external library calls to remain accurate, which we found to be a poor
engineering trade-off for our analytical workloads.
At termination, \texttt{submit\_answer} triggers an exporter that serialises
the claim graph together with $\mathcal{V}_{\mathrm{final}}$, recovers the
terminal evidence subgraph
$\mathcal{G}^{*}=\mathrm{Ancestors}_{\mathcal{G}}(\mathcal{V}_{\mathrm{final}})$
by backward closure over premise edges, and reads off the user-facing answer
$o$ as a deterministic function of $\mathcal{V}_{\mathrm{final}}$.
For QA tasks $|\mathcal{V}_{\mathrm{final}}|=1$ and $o$ is the content of the
single final claim. For report tasks $|\mathcal{V}_{\mathrm{final}}|>1$ and
$o$ is produced by the constrained writer $\mathrm{Compose}$ of
Eq.~\eqref{eq:terminal}, which reorders and concatenates
$\{\mathrm{content}(v)\}_{v\in\mathcal{V}_{\mathrm{final}}}$ into prose
without introducing new claim or data nodes. In both cases any factual
content visible to the user is attributable to a node of $\mathcal{G}^{*}$,
which is the invariant on which the traceability metrics of
Appendix~\ref{app:traceability-eval} rely. RL rollouts skip the report
writer and optimise directly against $\mathcal{V}_{\mathrm{final}}$ and
$\mathcal{G}^{*}$.

\subsection{Per-Turn Observation and Context Compression}
\label{app:verigraph-context}

\paragraph{Bounded rendering of $S_t$.}
The structured observation $S_t=(H_t,\mathcal{V}_t,\mathcal{G}_{t-1})$ defined
in Section~\ref{sec:primitives} cannot be passed verbatim, since
$\mathcal{V}_t$ may contain large dataframes and $\mathcal{G}_{t-1}$
accumulates over many turns. The runtime therefore exposes $S_t$ to the
policy through two deterministic summarisers:
\begin{itemize}
  \item $\phi_{\mathrm{ns}}(\mathcal{V}_t)$ reports each visible variable by
        type and shape (e.g., a \texttt{DataFrame} as $(\text{rows},\text{cols})$,
        an \texttt{ndarray} as $(\text{shape},\text{dtype})$, a scalar by a
        short preview), so the policy can address $v\in\mathcal{V}_t$ by name
        without paying for its content.
  \item $\phi_{\mathrm{cl}}(\mathcal{G}_{t-1})$ lists every $v\in\mathcal{V}_{\mathrm{final}}$
        already submitted, plus the most recent $K_{\mathrm{cl}}$ ($\!=\!30$)
        non-final claim nodes, each rendered as
        $(\text{id},\text{varname},\text{type},\mathrm{content}(v))$.
\end{itemize}
Each field is independently truncated to a per-field budget $b$
($\!=\!1{,}200$ characters) so that no single large object can exhaust the
prompt; \texttt{stdout} keeps its tail and \texttt{stderr} keeps its head,
matching the regions most useful for debugging.

\paragraph{Recent-$k$ history compression.}
Let $|\alpha_t|$ and $|z_t|$ denote the action and raw-observation token
counts at step $t$. The total context after $T$ turns is
$L_T=\sum_{t=1}^{T}(|\alpha_t|+|z_t|)$, and in data-analytic settings
$\mathbb{E}[|z_t|]\gg\mathbb{E}[|\alpha_t|]$ because $z_t$ may carry table
previews and tracebacks, so $L_T$ saturates the context window within only a
few turns if left uncompressed. The history component $H_t$ is therefore
maintained by a fixed-window compressor that keeps the newest $k$ tool
responses in full and replaces older ones with a constant stub:
\begin{equation}
  \tilde z_{t'} \;=\;
  \begin{cases}
    \textsc{Render}(z_{t'},\,\mathcal{V}_{t'},\,\mathcal{G}_{t'};\,b) & t' > t-k \\[2pt]
    \texttt{[omitted tool result]} & t' \le t-k
  \end{cases}
  \label{eq:compress}
\end{equation}
with $k\!=\!5$ in the reported configuration. Compression operates on whole
tool responses rather than on individual fields, preserving the alignment of
the $(\alpha_{t'},\tilde z_{t'})$ pairs that downstream loss-masking relies
on. Because persistent facts must be materialised as nodes of
$\mathcal{V}_{\mathrm{claim}}$ via \texttt{bind} or \texttt{infer}, dropping
older raw stdouts does not erase evidence: the relevant content has already
been promoted into $\mathcal{G}_{t-1}$ and re-enters every subsequent prompt
through $\phi_{\mathrm{cl}}(\mathcal{G}_{t-1})$.

\subsection{RL Integration}
\label{app:verigraph-rl-impl}

The same runtime is used for RL rollouts. Each RL prompt is paired with a
workspace directory and a judge-prompt type derived from the task family
(QA versus research), and rollouts are executed inside SGLang with a
per-trajectory copy of the executor. Algorithm~\ref{alg:verigraph} therefore
runs unchanged; the only differences are that (i)~the post-hoc report writer
is disabled, so the optimisation target is exactly $\mathcal{V}_{\mathrm{final}}$, and
(ii)~the rollout records, in addition to $\mathcal{G}^{*}$, a token-level loss
mask that selects assistant tokens generated by $\pi_\theta$ and excludes
\texttt{<tool\_response>} content.

\paragraph{Trajectory validity.}
A trajectory is marked valid if \texttt{submit\_answer} fired before the turn
budget $T$ and at least one tool call succeeded. Invalid trajectories
($\mathcal{V}_{\mathrm{final}}\!=\!\emptyset$ or zero successful tool calls)
have their loss mask zeroed, so the corresponding tokens contribute neither to
the policy gradient nor to the KL term. This protects the policy from
imitating its own malformed rollouts during early training.

\paragraph{Composite reward.}
RL uses the weighted reward detailed in Appendix~\ref{app:training}.

\subsection{Agent System Prompts Used at Training and Inference}
\label{app:verigraph-prompts}

For completeness and reproducibility, we list the exact system prompts that
the \ours policy receives during training and inference. The prompt in
Listing~\ref{lst:ours-inference-prompt} is the one used at \emph{evaluation
time} and during \emph{single-turn} rollouts (e.g., trajectory synthesis with
a strong teacher LLM); it spells out the workflow, the evidence-graph API,
the strict adherence rules, and few-shot working examples. The compact prompt
in Listing~\ref{lst:ours-sft-rl-prompt} is the system prompt that conditions
our 8B policy during \emph{SFT and RL training} as well as during multi-turn
inference: it preserves the same workflow and API but removes the few-shot
demonstrations, since at training/inference time the agent is already exposed
to canonical multi-turn trajectories. Finally,
Listing~\ref{lst:ours-report-prompt} is the system prompt fed to the
post-hoc report writer that turns $\mathcal{V}_{\mathrm{final}}$ into prose;
this writer is disabled during RL rollouts (Appendix~\ref{app:verigraph-rl-impl}).

\begin{promptbox}{\ours single-turn / inference system prompt (\texttt{VERIGRAPH\_PROMPT}). Few-shot examples are abbreviated.}{lst:ours-inference-prompt}
You are a rigorous Data Analysis Agent operating inside a persistent Python environment.

Your task is to conduct a comprehensive data analysis of the request and file provided by the user, and present all useful intermediate analysis conclusions in the form of claims. These claims will ultimately serve as material for report writing, so you need to provide as insightful and profound claims as possible.

### 1. Core Concept: The Claim System
A Claim is the only persistent unit of knowledge. Standard text you output is ephemeral and will be discarded by the system.
- Atomic Claims (bind): Direct, objective observations extracted from Python variables (e.g., "The mean weight is 42.5").
- Derived Claims (infer): Logical conclusions or judgments that synthesize multiple existing Claims.
- The Evidence Graph: Data -> bind() -> Atomic Claims -> infer() -> infer() ... -> Final Conclusion.

Each claim must be informative rather than a superficial description of the dataset; you need to conduct an in-depth analysis of the data.

### 2. Operational Workflow
You operate in a persistent Python environment. Follow this loop:
1. Analyze/Reasoning: use <think>...</think> to plan steps (loading, cleaning, analysis, synthesis) and write your internal reasoning.
2. Act: execute Python code in <code_interpreter>...</code_interpreter>. Whenever you want to express information or a conclusion, make claims in code. The environment is persistent; you can reuse variables from previous turns.
3. Observe: you will receive output in a <tool_response> block. Verify whether the Claim was successfully created and reflects the data.
4. Iterate: repeat the analyze-act-observe loop until you have sufficient claims.
5. Finish: call submit_answer() with the final claims. For QA, submit a single claim. For report generation, submit all claims needed by the report.

You have a total of 50 rounds of coding opportunities, so explore and analyze as much as needed.
For open-ended report tasks, plan first and then conduct multi-dimensional in-depth analysis (basic statistics as well as statistical methods).

### 3. Evidence Graph API
bind(template_str: str, **kwargs) -> Claim
- Ground objective facts from Python variables. Use {key} placeholders. Keep numeric values to 3 decimal places.
- No hardcoding: numbers must come from variables (e.g., bind("Mean is {m}", m=df['col'].mean())).
- Use this for user-provided facts: bind("User states X: {x}", x=user_input).
- Record only useful information; do not bind every inspection output.

infer(premises: list[Claim], conclusion: str, reasoning: str) -> Claim
- Use this to express reasoning in natural language when it cannot be expressed through code.
- premises must be a list of previously created Claim objects.
- conclusion must NOT introduce new raw data or facts not in premises; it must be supported by premises.
- The conclusion must NOT contain any placeholders.

submit_answer(final_claims: list[Claim] | Claim) -> None
- Finalize the task. Submit the final claim or a list of claims (e.g., [c1, c2, c3_conclusion]).
- Before submit_answer, confirm that the claims you use meet your expectations (see claim content from system logs).

### 4. Strict Adherence Rules
- Zero-Text Policy: do not provide answers in plain text; important insights must be Claims.
- Persistence Awareness: reuse variables, imports, DataFrames; do not re-load files unless necessary.
- Traceability: before submit_answer, ensure a human can reconstruct your logic using only the Claims you created.
- Verification: never infer based on a Claim you have not seen successfully created in the tool response.
- Every time you use infer or submit_answer, first confirm the claims you use meet your expectations.
- Do not create duplicate Claims for the same fact; reuse existing Claim variables as premises when possible.
- Context Management: tool outputs are truncated and older outputs may be dropped. Never rely on old stdout for important facts; bind them into Claims.

### 5. How To Run Code
Write a Python block exactly like:
<code_interpreter>
# valid python only
</code_interpreter>
Place bind() / infer() / submit_answer() calls inside that Python block.

Safety: do not write or delete files, run shell commands, or make network requests. Keep prints short and only print what you need to decide the next step.

### 6. Working Examples
[Example 1: Open-Ended Research]
User: Analyze 'nhl_data.csv' for player longevity and physical traits.
Assistant: <think> plan inspection, baseline statistics, correlation between physical traits and Total_Games, comparative synthesis ... </think>
<code_interpreter>
import pandas as pd
df = pd.read_csv("nhl_data.csv")
n_players = df['player_id'].nunique()
h_stats = df['height'].describe(); w_stats = df['weight'].describe()
c1 = bind("The dataset tracks {n} unique players. Heights {h_min}-{h_max} cm (avg {h_avg:.1f}); weights {w_min}-{w_max} kg (avg {w_avg:.1f}).",
          n=n_players, h_min=h_stats['min'], h_max=h_stats['max'], h_avg=h_stats['mean'],
          w_min=w_stats['min'], w_max=w_stats['max'], w_avg=w_stats['mean'])
</code_interpreter>
<tool_response> Successfully created claim ... </tool_response>
... [further bind / infer steps elided] ...
<code_interpreter>
c4 = infer(premises=[c1, c2, c3],
           conclusion="In professional hockey, body mass is a better predictor of career longevity than height.",
           reasoning="Height correlates weakly (0.12) while weight correlates moderately (0.45); the heavy/light gap is ~150 games.")
submit_answer([c1, c2, c3, c4])
</code_interpreter>

[Example 2: Short QA]
User: What is the average BMI of patients in 'patients.csv'?
Assistant: load file, compute BMI = weight / (height/100)^2, take the mean, then
<code_interpreter>
c_bmi = bind("The average BMI across the patient cohort is {avg:.2f} kg/m^2.", avg=avg_bmi)
submit_answer(c_bmi)
</code_interpreter>
\end{promptbox}

\begin{promptbox}{\ours SFT/RL training and multi-turn inference system prompt (\texttt{VERIGRAPH\_PROMPT\_SFT}).}{lst:ours-sft-rl-prompt}
You are a rigorous Data Analysis Agent operating in a persistent Python environment. Your task is to analyze data and formulate insights strictly using a Claim-based API.

### 1. Workflow
1. Plan: use <think>...</think> to outline your reasoning and next steps.
2. Execute: write Python code inside <code_interpreter>...</code_interpreter>.
3. Observe: wait for the system's <tool_response>.
4. Submit: once you have sufficient evidence, call submit_answer().

### 2. Evidence Graph API (Available in Python)
You MUST construct all insights using these persistent Claim objects:
- bind(template_str: str, **kwargs) -> Claim: create an atomic claim directly from data variables. Format numbers to 3 decimal places. Example: c1 = bind("Mean is {m}", m=val).
- infer(premises: list[Claim], conclusion: str, reasoning: str) -> Claim: derive a logical conclusion from existing claims; do not introduce new raw data here.
- submit_answer(final_claims: list[Claim] | Claim) -> None: terminate the task and submit your final claims as the answer.

### 3. Strict Constraints
- Zero-Text Policy: standard text outside code blocks is ephemeral. All meaningful insights and final answers must be instantiated via the Claim APIs.
- Persistence: the Python environment is persistent. Reuse imported libraries, DataFrames, and previously created Claim variables.
- Verification: never pass a Claim to infer or submit_answer until you have verified its successful creation in the preceding <tool_response>.
\end{promptbox}

\begin{promptbox}{Post-hoc report writer prompt (\texttt{REPORT\_PROMPT}). Disabled during RL rollouts.}{lst:ours-report-prompt}
You are given a user question and a list of final claims.
Write a clean final report in Markdown using only those claims.
Do not invent facts.
\end{promptbox}

% =====================================================================

\section{Training and Optimization Details}
\label{app:impl-details}

\subsection{Training Data Construction Details}
\label{app:data}

We curate the training corpus from a heterogeneous pool of publicly available
table-grounded and data-analysis benchmarks, and re-purpose them into evidence-graph
trajectories that exercise the primitives defined in Section~\ref{sec:method}.
Throughout this section we denote a trajectory by $\tau = (q, \{(a_t, o_t)\}_{t=1}^{T})$,
where $q$ is the task with its associated files, $a_t$ a graph-aware action, and $o_t$
the corresponding observation.

\paragraph{Source datasets.}
Our training corpus follows the public datasets and data mixtures used in prior
table-reasoning and data-agent work~\cite{deepanalyze,datamind}, rather than introducing new task instances or additional benchmark annotations. We draw on six upstream source families that cover complementary regimes of
table-grounded and data-analytic reasoning. The pool consists of TableInstruct,
the instruction data released with TableBench~\cite{tablellm} ($\sim$19K);
TAT-QA~\cite{tatqa} ($\sim$13K), a finance QA benchmark over hybrid tabular and
textual evidence; CRT-QA~\cite{crtqa} ($0.7$K), which emphasizes complex
reasoning over tables; MultiHiertt~\cite{multihiertt} ($\sim$7K), which requires
numerical reasoning over multi-hierarchical tables and associated text;
DataScience-Instruct from DeepAnalyze~\cite{deepanalyze}, split into a TableQA
partition ($\sim$3K) and an open-ended data-analysis partition ($\sim$10K); and
DataMind-54K~\cite{datamind} ($\sim$7.3K). Together, these sources span
single-table QA, hybrid table--text QA, multi-table and multi-document reasoning,
and open-ended data analysis. The single-table QA sources provide short,
high-precision supervision for atomic evidence-graph operations, whereas
MultiHiertt, DataMind-54K, and the open-ended
DataScience-Instruct partition contribute longer trajectories that require
cross-file computation, quantitative synthesis, and evidence-preserving report
generation.

\subsubsection{SFT Data}
\label{app:sft-data}

The SFT data is constructed to teach the model both how to solve data-analysis
tasks and how to maintain the evidence graph while solving them. We first
synthesize complete task trajectories from the source datasets, then filter them
for executable and logically consistent graph construction, and finally derive
atomic next-action samples from the retained trajectories.

\paragraph{Trajectory synthesis.}
For each source example, we use a stronger teacher model (Qwen3-32B) to generate a full solution trajectory. The teacher is run through the same \ours runtime rather than asked to emit a static transcript:
it receives the system prompt in Listing~\ref{lst:sft-trajectory-prompt}, calls
the Python sandbox with the evidence-graph API, observes the executor feedback,
and stops only after calling \texttt{submit\_answer}. The annotation script uses
temperature $0.6$, top-$p$ $0.95$, top-$k$ $20$, repetition penalty $1.1$, a
maximum of $8{,}192$ generated tokens per model call, $120$s per code cell, and
$400$s per task. We cap each trajectory at 50 interaction turns to avoid
unbounded generation while still allowing long multi-step analyses.

\begin{promptbox}{Trajectory annotation system prompt used for SFT synthesis. Demonstration examples in the code are omitted for space.}{lst:sft-trajectory-prompt}
You are a rigorous Data Analysis Agent operating inside a persistent Python environment.

Your task is to conduct a comprehensive data analysis of the request and file provided by the user, and present all useful intermediate analysis conclusions in the form of claims.
These claims will ultimately serve as material for report writing, so you need to provide as insightful and profound claims as possible.

### Core Concept: The Claim System
A Claim is the only persistent unit of knowledge. Standard text you output is ephemeral and will be discarded by the system.
- Atomic Claims (bind): Direct, objective observations extracted from Python variables.
- Derived Claims (infer): Logical conclusions or judgments that synthesize multiple existing Claims.
- The Evidence Graph: Data -> bind() -> Atomic Claims -> infer() -> ... -> Final Conclusion.

### Operational Workflow
1. Use <think>...</think> to plan loading, cleaning, analysis, and synthesis.
2. Execute Python code inside <code_interpreter>...</code_interpreter>.
3. Whenever you want to express information or a conclusion, create a claim in Python.
4. Review the <tool_response> to verify whether the claim was successfully created.
5. Repeat until there are sufficient claims, then call submit_answer().

### Evidence Graph API
bind(template_str: str, **kwargs) -> Claim
- Ground objective facts from Python variables. Keep numeric values to 3 decimals.
- Do not hardcode numbers; values must come from variables.

infer(premises: list[Claim], conclusion: str, reasoning: str) -> Claim
- Derive a natural-language conclusion from existing claims.
- Do not introduce raw data or facts that are not supported by the premises.

submit_answer(final_claims: list[Claim] | Claim) -> None
- Finalize the task. For QA, submit the final claim; for report tasks, submit all claims needed by the report.

### Strict Rules
- Zero-Text Policy: all meaningful insights and final answers must be instantiated as Claims.
- Persistence: reuse variables, imports, DataFrames, and existing Claim variables.
- Traceability: before submit_answer, ensure a human can reconstruct the logic using only the created Claims.
- Verification: never infer from or submit a Claim until its successful creation appears in the tool response.
- Safety: do not write files, delete files, run shell commands, or make network requests.
\end{promptbox}

% yztodo: 细化一下这里的筛选以及代码的atomic sample的细节，比如最后构造了哪些原子数据。
\paragraph{Trajectory filtering.}
% Because synthesized trajectories may contain malformed code, invalid tool-call
% syntax, evidence-graph violations, or inconsistent reasoning, we apply automatic
% filters followed by targeted manual inspection. The filtering pipeline removes:
% (i) trajectories with invalid formats, including malformed tool calls, missing
% graph-maintenance actions, or no final submitted answer; (ii) trajectories whose
% code execution is unreliable, defined as more than $50\%$ of code cells failing;
% (iii) trajectories shorter than 256 tokens or longer than 32K tokens; and
% (iv) trajectories whose final answers receive a low LLM-as-judge score when
% compared against the gold answer or task-specific rubric. This leaves only
% trajectories that are executable, structurally valid, and aligned with the
% expected answer.
Because synthesized trajectories may contain malformed code, invalid tool-call syntax, evidence-graph violations, or inconsistent reasoning, we apply automatic filters followed by targeted manual inspection. The filtering pipeline removes trajectories that are unsuitable for training, including those with invalid formats, malformed tool calls, missing graph-maintenance actions, no final submitted answer, or unreliable code execution, defined as more than 50\% of code cells failing. We also discard trajectories shorter than 256 tokens or longer than 32K tokens, as well as trajectories whose final answers receive a low LLM-as-judge score when compared against the gold answer or task-specific rubric. Rather than retaining only error-free traces, we preserve a small fraction of executable and structurally valid trajectories that contain localized mistakes, failed intermediate attempts, or suboptimal reasoning steps, provided that they are ultimately recoverable and lead to an acceptable final answer. We define recoverable errors as localized mistakes in intermediate steps that do not invalidate the trajectory structure or final answer, and that are corrected or safely bypassed in subsequent steps. This yields a training set that remains structurally reliable and aligned with the expected answer, while exposing the model to controlled error-and-recovery patterns that may arise during problem solving.

\paragraph{Atomic sample construction.}
% From the filtered full trajectories, we construct fine-grained SFT samples with a
% sliding-window scheme. Given a trajectory with $T$ turns, we keep the prefix up to
% turn $t$ as context and use the next action at turn $t+1$ as the prediction target.
% Repeating this over each trajectory yields multiple atomic training examples, each
% focused on the next evidence-graph operation or tool action. We keep three atomic
% task types: first-step planning from the user query and file list, next-action
% prediction after compressed tool feedback, and final submission / report
% construction from the completed claim set. These samples complement
% full-trajectory imitation by directly supervising local graph-maintenance
% decisions.
From the filtered full trajectories, we construct fine-grained SFT samples with a sliding-window scheme. Given a trajectory with $T$ turns, we keep the prefix up to turn $t$ as context and use the next action at turn $t+1$ as the prediction target. To control context length and remove redundant observations, the per-turn context is compressed before constructing atomic samples, following the procedure described in \hyperref[app:verigraph-context]{Appendix~C.3}. Repeating this over each trajectory yields multiple atomic training examples, each focused on the next evidence-graph operation or tool action. During atomic sample construction, we retain an example only when the target next action and its resulting feedback pass the validity and correctness checks; if the execution result of the target step is erroneous, the corresponding atomic example is discarded. However, the preceding context is not required to be error-free. When earlier turns contain localized mistakes but the current target step correctly recovers from, corrects, or bypasses them, we preserve those imperfect historical contexts. This allows the model to learn not only standard next-action prediction under clean contexts, but also how to perform self-correction in the presence of prior errors. We keep three atomic task types: first-step planning from the user query and file list, next-action prediction after compressed tool feedback, and final submission/report construction from the completed claim set. These samples complement full-trajectory imitation by directly supervising local graph-maintenance decisions and recovery-oriented actions.

\begin{table}[htbp]
\centering
\caption{SFT and RL data mixture. Counts are approximate.}
\label{tab:app-sft-mixture}
\footnotesize
\begin{tabular}{@{}lrr@{}}
\toprule
\textbf{Source} & \textbf{SFT} & \textbf{RL} \\
\midrule
DataScience-Instruct (TableQA)  & 2.6K & 2.0K \\
DataScience-Instruct (open-end) & 0.7K & 3.7K \\
CRT-QA                          & 0.8K & 0.2K \\
MultiHiertt                     & 4.3K & 0.7K \\
TableInstruct                   & 15.6K & 2.0K \\
TAT-QA                          & 11.0K & 1.3K \\
DataMind                        & 6.5K & 2.1K \\
\midrule
\textbf{Total}                  & \textbf{42K} & \textbf{12K} \\
\bottomrule
\end{tabular}
\end{table}

% yztodo: table2的数据量check一下，然后rl部分的构造过程也补充一下。
\subsubsection{RL Data}
\label{app:rl-data}

RL prompts are drawn from the same pool as the SFT corpus but with strictly
non-overlapping queries; the per-source quotas are listed in
Table~\ref{tab:app-sft-mixture}. Short single-table QA sources are down-weighted
and long multi-document analysis sources are up-weighted, so that the RL stage
focuses on prompts that genuinely benefit from multi-step graph construction.

\paragraph{Difficulty filtering.}
To avoid the well-known reward-collapse phenomenon on prompts that are either
saturated or unsolvable, we run $K\!=\!8$ rollouts per prompt with the SFT-only
policy, estimate the empirical pass rate $\hat{p}$, and keep prompts with
$\hat{p} \!\in\! [p_{\min}, p_{\max}]\!=\![0.1, 0.8]$. For open-ended tasks, the
``pass'' label is replaced by a rubric score above threshold from the same
LLM-as-judge used in SFT filtering. The classifier prompt used for difficulty
annotation is shown in Listing~\ref{lst:rl-difficulty-prompt}.

\begin{promptbox}{RL difficulty-classifier prompt (abbreviated).}{lst:rl-difficulty-prompt}
Given a task description, the data files, K rollouts of the SFT policy, and the
gold answer (or rubric), label the task as {trivial, easy, medium, hard,
unsolvable} based on the empirical pass rate and reasoning depth required.
Output JSON: {"pass_rate":..., "depth":..., "label":..., "keep":true|false}.
Keep iff label in {easy, medium, hard} and pass_rate in [0.1, 0.8].
\end{promptbox}

\paragraph{Final RL pool.} The above procedure produces $12$K RL prompts. For fast
hyper-parameter sweeps we additionally use a stratified $1$K subset that preserves
the per-source proportions of the full pool.

\subsection{Training Details}
\label{app:training}

\paragraph{SFT.}
We train the Qwen3-8B backbone with full-parameter SFT using MS-Swift~\cite{msswift}.
The corpus contains the 42K examples in Table~\ref{tab:app-sft-mixture}. We first
train on the atomic next-action samples and then mix in complete trajectories so
that the model learns both the primitive syntax and long-horizon graph
construction. Table~\ref{tab:app-sft-hparams} lists the implementation settings.

\begin{table*}[t] % 跨双栏通常放在页面顶部 [t]
\centering
\caption{Key hyperparameters in SFT phase.}
\label{tab:app-sft-hparams}
\small % 稍微调大一点，footnotesize 有时太小
\begin{tabular}{lc} % 使用 l (左对齐) 和 c (居中)
\toprule
\textbf{Hyperparameter} & \textbf{Value} \\
\midrule
Max Sequence length & 32768 \\
Training Epochs & 2 \\
Effective global batch size & 128 \\
Per-device train / eval batch & 2 / 4 \\
Gradient accumulation & 16 \\
Learning rate & $1 \times 10^{-5}$ \\
Warm-up ratio & 0.05 \\
Precision & bfloat16 \\
\bottomrule
\end{tabular}
\end{table*}

\paragraph{RL.}
We initialize RL from the SFT checkpoint and optimize with DAPO~\cite{dapo} in
Verl~\cite{verl}. The rollout engine is SGLang in multi-turn mode, using the
same \ours code executor and evidence-graph runtime as evaluation. Prompts are
converted to chat messages with the SFT-time VeriGraph prompt, a user question,
and the attached file names; the preprocessor keeps the task directory in
\texttt{extra\_info} so each rollout executes in the correct workspace. Table~\ref{tab:app-rl-hparams} reports the main composite-reward setting;
the fast ablation scripts reduce the rollout group size to 4.

\begin{table*}[t]
\centering
\caption{Key hyperparameters in RL phase.}
\label{tab:app-rl-hparams}
\small
\begin{tabular}{llc}
\toprule
\textbf{Category} & \textbf{Hyperparameter} & \textbf{Value} \\
\midrule
\textbf{Training} & Rollout Backend & SGLang \\
& Rollouts per prompt & 8 \\
& Train prompt batch size & 16 \\
& Generation prompt batch size & 4 \\
& PPO mini-batch size & 2 \\
& PPO micro-batch per GPU & 1 \\
& Learning rate & $1\times10^{-6}$ \\
& LR warm-up steps & 10 \\
& Weight decay & 0.1 \\
& KL loss coefficient & 0.01 \\
\midrule
\textbf{Rollout} & Temperature & 1.0 \\
& Top-$p$ & 0.95 \\
& Repetition penalty & 1.1 \\
& Max response length & 8192 \\
& Max interaction turns & 50 \\
& Tool observation budget & 1,200 chars \\
& Tool timeout & 120s \\
& Trajectory timeout & 1,800s \\
\midrule
\textbf{Reward} 
& Process reward weight & 0.40 \\
& Final reward weight & 0.35 \\
& Infer reward weight & 0.15 \\
& Reward clipping (Min) & $-0.3$ \\
& Reward clipping (Max) & $0.8$ \\
& Reward judge model & \texttt{gpt-4o-mini} \\
\bottomrule
\end{tabular}
\end{table*}

\paragraph{Reward design.}
The reward shaping signal in our RL loop is produced by a unified verifier
module that decomposes the trajectory-level reward into the same three
components introduced in the main text (Eq.~\ref{eq:reward}), namely
$R_{\text{process}}$, $R_{\text{infer}}$, and $R_{\text{outcome}}$, each
targeting a distinct failure mode of long-horizon analytic reasoning. To
stabilise optimisation in practice, the implemented trajectory reward
extends the main-text composite with per-component weights, a
missing-submission penalty, and an overlong-response shaping term:
\begin{equation}
\label{eq:app-composite-reward}
R \;=\; w_{\text{p}}\,R_{\text{process}} \;+\; w_{\text{i}}\,R_{\text{infer}} \;+\; w_{\text{f}}\,R_{\text{outcome}}
       \;+\; \mathbb{1}[\,\mathcal{V}_{\mathrm{final}}\!=\!\emptyset\,]\cdot p_{\text{sub}} \;+\; R_{\text{overlong}},
\end{equation}
where $\mathcal{V}_{\mathrm{final}}$ denotes the final claim set submitted by
the policy. The weights $(w_{\text{p}},w_{\text{i}},w_{\text{f}})$ and the
missing-submission penalty $p_{\text{sub}}$ are taken from
Table~\ref{tab:app-rl-hparams}, $R_{\text{overlong}}$ is inherited from
DAPO~\cite{dapo}, and a final clipping step to $[-0.3, 0.8]$ is applied for
variance control. Setting $w_{\text{p}}=w_{\text{i}}=w_{\text{f}}=1$ and
disabling the auxiliary terms recovers the main-text form exactly.
Following standard practice for trajectory-level RL with token-level loss
masking, the scalar reward is written onto the last assistant-generated token,
so that all generated tokens share a single advantage during the policy update.
The infer-edge and outcome verifiers are realised by the same off-the-shelf judge model
(\texttt{gpt-4o-mini} in our experiments), and judge calls are issued
asynchronously across the rollout batch under a fixed concurrency budget so as
to avoid stalling the rollout--update pipeline.

\emph{Process reward.}
$R_{\text{process}}$ is a deterministic, verifier-free signal that captures
the syntactic and operational health of the trajectory. Following
Eq.~\ref{eq:process_reward} in the main text, it is the empirical success
rate of the executed actions against the Python sandbox,
\[
R_{\text{process}} \;=\; \frac{1}{T}\sum_{t=1}^{T}\mathbb{I}[\texttt{exec}(\alpha_t)=\text{success}],
\]
and set to zero whenever the trajectory is marked invalid or contains no tool
calls. Because this term is computed from the executor's own feedback, it
incurs no additional verifier traffic and can be evaluated densely at no cost.
Empirically it serves to stabilise early training, where the dominant failure
mode is malformed code rather than substantive reasoning errors.

\emph{Inference reward.}
$R_{\text{infer}}$ is the central novelty of our reward design and operates
at the granularity of individual derivational edges in the evidence graph.
For each edge $(\mathcal{P}_i, v_{\mathrm{new}})\in\mathcal{E}_{\mathrm{derive}}$
introduced by an \texttt{infer} primitive
(Eq.~\ref{eq:derive_expansion}), the verifier receives the rendered contents
of the premise claims $\mathcal{P}_i$, the conclusion $c_i = \mathrm{content}(v_{\mathrm{new}})$,
and the natural-language justification $r_i = \mathrm{reasoning}(v_{\mathrm{new}})$,
and is asked to decide whether $c_i$ is logically entailed by, or at
least clearly licensed by, $\mathcal{P}_i$. Concretely, this realises
$\texttt{Verify}(q, \mathcal{P}_i, r_i, c_i)\in\{-1,+1\}$ in
Eq.~\ref{eq:infer_reward}: $+1$ for an inference that is faithful to its
premises and $-1$ for one that introduces an unsupported leap, a
hallucinated quantity, a speculative causal attribution, or a contradiction.
The per-trajectory inference reward is the
mean of these scores over all derivational edges in $\mathcal{G}^{*}$, and is
defined to be zero when no \texttt{infer} edge is present. The exact judge
prompt is given in Listing~\ref{lst:rl-infer-judge}.
Crucially, this term penalises locally invalid reasoning even when the final
answer happens to be numerically correct, and thereby couples optimisation
pressure to the structural integrity of the evidence graph rather than to its
terminal output alone.

\emph{Outcome reward.}
$R_{\text{outcome}}$ instantiates the main-text definition
(Eq.~\ref{eq:outcome_reward}),
$R_{\text{outcome}} = \mathbb{I}[\text{terminal extraction}]\cdot\texttt{Judge}(q,\mathcal{G}^{*},a^*)/S$,
and measures the quality of the submitted answer relative
to the task specification. It is computed only for valid trajectories that
terminate via \texttt{submit\_answer} with a non-empty
$\mathcal{V}_{\mathrm{final}}$; trajectories that fail to submit incur the
fixed penalty $p_{\text{sub}}$ in Eq.~\ref{eq:app-composite-reward} so that
the verifier is never invoked on degenerate rollouts. To accommodate the
heterogeneity of the RL pool, the outcome verifier dispatches between two
task-conditioned judge prompts. For closed-form analytic tasks with a gold
answer, the verifier evaluates $\mathcal{V}_{\mathrm{final}}$ against the
reference along the axes of accuracy, completeness, and hallucination
(Listing~\ref{lst:rl-qa-judge}). For open-ended research tasks for which no
gold answer exists, the verifier instead scores the relevance, analytical
depth, and logical soundness of the submitted claims against the user request,
conditioning additionally on a compact summary of the rollout evidence so that
the score reflects whether the claims are supported by what the agent actually
observed rather than by external priors of the judge
(Listing~\ref{lst:rl-research-judge}). In both cases the judge is required to
emit a structured JSON object with an ordinal score
$s = \texttt{Judge}(q,\mathcal{G}^{*},a^*)\in\{0,1,2,3\}$ accompanied by a
brief justification; with maximum rubric score $S=3$ this yields
$R_{\text{outcome}} = s/S \in [0,1]$, and a
malformed or unparseable judge response is treated as $s=0$.

\begin{promptbox}{Outcome verifier system prompt for closed-form QA tasks.}{lst:rl-qa-judge}
You are an expert data analysis evaluator. Your task is to evaluate the quality of a set of analytical findings (claims) generated by an AI agent in response to a specific question. You will be provided with the user's question, the ground truth (golden answer), and the agent's generated claims.

Please evaluate the agent's claims based on how well they align with the golden answer.

### Evaluation Criteria:
1. Accuracy: Do the claims state facts that are correct according to the golden answer?
2. Completeness: Did the agent capture the core message and key points of the golden answer?
3. Hallucination: Did the agent invent data points, trends, or facts that contradict or are unsupported by the golden answer?

### Scoring Rubric (0-3):
* [3] Excellent (Fully Accurate & Complete): The claims perfectly capture the key points of the golden answer. There are no factual errors, no severe omissions, and zero hallucinations.
* [2] Partial (Minor Errors or Omissions): The claims capture the core of the golden answer but miss some minor details, or include slight, harmless extra information. No severe hallucinations or direct contradictions.
* [1] Poor (Severe Errors or Hallucinations): The claims contain significant factual errors, contradict the golden answer, miss the primary core message, or suffer from major hallucinations.
* [0] Invalid: The response is entirely empty, nonsensical, or completely unrelated to the question.

### Input Data:
[Question]: {question}
[Golden Answer]: {golden_answer}
[Agent Output]: {agent_claims}

### Output Format:
You must output your evaluation in the following JSON format. Provide a concise step-by-step reasoning BEFORE giving the final integer score.
{"reasoning": "Briefly compare the claims against the golden answer based on accuracy, completeness, and hallucination.", "score": <int between 0 and 3>}
\end{promptbox}

\begin{promptbox}{Outcome verifier system prompt for open-ended research tasks.}{lst:rl-research-judge}
You are a Senior Data Scientist and an expert evaluator. Your task is to evaluate the quality of exploratory data analysis findings (claims) generated by an AI agent for an open-ended research topic. There is no predefined correct answer.

Please evaluate the agent's claims based on their analytical value, logical soundness, and relevance to the research topic.

### Evaluation Criteria:
1. Relevance: Do the claims directly address the core intent of the research question?
2. Depth & Insightfulness: Do the claims demonstrate deep analytical thinking (e.g., correlations, anomalies), or are they just superficial observations?
3. Logical Soundness: Are the conclusions drawn plausible and logically supported by standard data analysis practices?

### Scoring Rubric (0-3):
* [3] Excellent (Highly Insightful & Logical): Highly relevant and logically sound. The claims go beyond obvious, surface-level observations to provide valuable, deep analytical perspectives.
* [2] Fair (Relevant but Superficial): The claims are relevant and generally logical, but lack analytical depth. They mostly state obvious facts or simple summaries without extracting deeper insights.
* [1] Poor (Flawed or Irrelevant): The claims draw highly questionable or illogical conclusions, lack structural coherence, or fail to meaningfully address the core research topic.
* [0] Invalid: The response is entirely empty, nonsensical, or completely ignores the research topic.

The user prompt additionally appends a compressed multi-turn evidence summary
(at most 6 messages, 280 chars each) so that the judge can verify whether the
final claims are supported by what actually happened during the rollout.

### Output Format:
{"reasoning": "Briefly evaluate the claims focusing on relevance, analytical depth, and logical soundness.", "score": <int between 0 and 3>}
\end{promptbox}

\begin{promptbox}{Infer verifier system prompt, applied once per \texttt{infer} edge.}{lst:rl-infer-judge}
You are an expert evaluator for natural language inference in a claim-based data analysis workflow.

Your task is to evaluate whether a single infer step is logically valid based on its premises.

You will be given:
1. The original user question
2. A set of premise claims
3. The inferred conclusion
4. The reasoning text

### Scoring Rules
* [1] Good inference
  - The conclusion is supported by the premises.
  - The reasoning is logically sound and faithful to the premises.
  - No unsupported leap, hallucination, contradiction, or speculative extension is introduced.
* [-1] Bad inference
  - The conclusion is not supported by the premises, or overstates what the premises justify.
  - The reasoning introduces unsupported assumptions, speculative causal interpretation, hidden leaps, contradiction, or distortion.

### Important Instructions
- Judge ONLY whether this infer step itself is valid from the premises.
- Do NOT judge based on whether the final answer is correct.
- If the conclusion adds new causal/mechanistic/speculative interpretation not grounded in the premises, assign -1.
- Be conservative: assign 1 only when the inference is clearly justified by the premises.

### Output Format
{"reasoning": "Briefly explain whether the conclusion is supported by the premises and whether there is any unsupported leap.", "score": 1 or -1}
\end{promptbox}

\paragraph{Reward-Model Overhead.}
The composite reward adds verifier traffic on top of standard outcome-only RL only through $R_{\text{infer}}$ (one verifier query per \texttt{infer} edge) and $R_{\text{outcome}}$ (one judge query per rollout); $R_{\text{process}}$ is computed locally from execution feedback at no extra cost. Per task tuple this incurs $\mathcal{O}\bigl(N \cdot (|\mathcal{I}| + 1)\bigr)$ verifier calls, where $|\mathcal{I}|$ is empirically bounded by the number of derivational steps in a rollout (typically $\leq 10$). We batch verifier calls across rollouts and use a small dedicated verifier rather than the policy itself, so the verifier is not on the critical path of the policy update; end-to-end this keeps reward-model overhead within roughly $1.3{\times}$ the rollout cost of an outcome-only baseline at the same $N$.

\section{Experimental Protocol}
\label{app:eval}

\subsection{Benchmark Details}

\paragraph{Datasets.}
We evaluate \ours on four data-intensive benchmarks that collectively span single-table QA, multi-table data analysis, and multi-step research across heterogeneous sources. For all benchmarks, we utilize the official released splits without any further re-annotation. Specific details for each benchmark are delineated below:
\begin{itemize}
\item \textbf{TableBench}~\cite{tablellm} (\emph{Table QA}): TableBench is a comprehensive single-table question-answering benchmark encompassing four distinct reasoning skills: fact checking, numerical reasoning, data analysis, and visualization. We utilize the public test split and evaluate the $\sim$700 textual questions from the fact-checking, numerical-reasoning, and data-analysis subsets; the visualization subset is excluded, as it requires plot-based grading that is orthogonal to our focus on claim-grounding evaluation. Each instance comprises a single CSV or Markdown table paired with a natural-language question, where the expected answer is a short string, a numerical value, or a list.
\item \textbf{InfiAgent-DABench}~\cite{infiagentdabench} (\emph{Data Analysis}): InfiAgent-DABench targets closed-form data analysis on individual CSV files. We employ the official evaluation split consisting of $257$ questions; each question specifies an analysis task—such as descriptive statistics, filtering-then-aggregation, or group-wise comparison—alongside an output format constraint. The gold answers are concrete values or short tuples; models are required to load the CSV, execute the analysis, and generate an answer that strictly adheres to the specified format.
\item \textbf{DSBench}~\cite{dsbench} (\emph{Multi-table/Long-context Data Analysis}): DSBench evaluates agents on realistic data-science workflows involving multiple related tables and supplementary documents. We focus on the data-analysis track and evaluate $466$ tasks; each task provides a bundle of CSVs—often containing hundreds of columns and tens of thousands of rows—plus a task description. The goal is to provide a numerical or short-text answer that necessitates joining, filtering, and aggregating across multiple tables. Given that the inputs frequently exceed standard context windows, DSBench specifically tests an agent's capability to navigate large workspaces through code execution.
\item \textbf{DAB-Step Research} (\emph{Multi-step Research}): DAB-Step~\cite{dabstep} is a multi-step benchmark requiring joint reasoning over structured tables and unstructured policy or documentation files within a payments-processing context. While the original benchmark focused on simple QA, we evaluate on the 100-case subset released by DeepAnalyze~\cite{deepanalyze}, adopting their official split for direct comparability. Each instance includes a question demanding cross-source reasoning, the associated tabular inputs, and relevant documentation snippets. The outputs are open-ended research-style responses rather than short answers; following~\cite{deepanalyze}, an LLM judge scores each response on \emph{Content} (factual correctness and completeness relative to the gold reference) and \emph{Format} (structure, citation discipline, and presentation) on a $0$--$5$ scale, and we report the average scores.
\end{itemize}

\noindent\textbf{Grounding Rate (GR).} On all four benchmarks we additionally
report the Grounding Rate defined in Appendix~\ref{app:traceability-eval}.
GR is computed from the model's own output and evidence context and therefore
does not depend on dataset-specific gold annotations, which makes it directly
comparable across the four benchmarks above.

\subsection{Baseline Details}
\label{app:baselines}
We group baselines into three categories. All baselines receive the same user
question and the same attached files. For QA-style benchmarks we extract the
content inside \texttt{<answer>...</answer>} when present; for report-style tasks
we pass the full generated report to the judge.

\paragraph{(i) General-purpose LLMs, direct prompting.}
The direct-inference baselines in Table~\ref{tab:main_acc} are GPT-5.2,
Gemini-2.5-Pro, Claude-4.5-Sonnet, Claude-4.5-Opus, Qwen3-32B, and
Qwen3-30B-A3B. They do not receive a Python tool. The evaluator renders each
attached file into text before prompting: text-like files are read directly,
spreadsheet / parquet / pickle files are converted through pandas previews, and
the input is truncated to at most 20K characters per file and 60K characters in
total. The exact system and user prompt template is shown in
Listing~\ref{lst:bl-direct-prompt}.

\begin{promptbox}{Direct-inference baseline prompt template.}{lst:bl-direct-prompt}
System:
You are a rigorous data analysis assistant.

You will receive a user question plus the textual contents of any attached files.
Files may be tables, HTML pages, JSON, Markdown, logs, source code, or other text-like data.

Your task:
1. Use only the provided question and file contents.
2. Inspect the file contents carefully before answering.
3. If the attached content is tabular, reason from the rows, columns, headers, and values shown.
4. If the attached content is HTML, reason from the visible text and relevant structured markup.
5. If the evidence is incomplete or truncated, say what can be answered from the visible content and avoid inventing missing facts.
6. Provide the final answer inside <answer></answer> tags. For report-style requests, put the full report inside those tags.

User:
Question:
{question}

Attached file contents:
{rendered_file_block}

Answer the question using the provided content. Put the final answer inside <answer></answer> tags.
\end{promptbox}

\paragraph{(ii) ReAct / CodeAct data agents.}
The ReAct-style baselines are GPT-5.2, GPT-5.4, Claude-4.5-Sonnet,
Claude-4.5-Opus, Qwen3-8B, Qwen3-Coder-30B, QwQ-32B, Qwen3-32B, and
Qwen3-30B-A3B. Each model is wrapped by the same CodeAct scaffold: it can issue
Python code in \texttt{<code\_interpreter>} blocks, receives executor output in
a tool-response block, and must finish with \texttt{<answer>...</answer>}. The
maximum interaction budget is 30 Python calls, with a 1,200-character cap on
each tool observation. The prompt in Listing~\ref{lst:bl-react-prompt} is the
system prompt used by the scaffold; the implementation includes an additional
one-shot height/weight correlation example after these instructions.

\begin{promptbox}{ReAct / CodeAct baseline system prompt.}{lst:bl-react-prompt}
You are a rigorous Data Analysis Agent operating within a persistent Python environment. Your core objective is to solve user's question through iterative code execution and produce a final answer.

## Iterative Workflow
1. Analyze: Write your internal reasoning process inside <think> tags.
2. Act: Execute Python code using the <code_interpreter> format.
3. Observe: You will receive the output in a <tool_response> block. Repeat until you have sufficient data.
4. Answer: Provide your final answer inside <answer></answer> tags. Whether the answer is a specific small number or a report, enclose it with <answer> and </answer> tags.

Before conducting the analysis, you should first check the data and the question to understand the user's intent.

## Tool Usage Format
To execute code, strictly use the following tag structure.
<code_interpreter>
# Write valid Python code here
import pandas as pd
print("Hello World")
</code_interpreter>

## Python Environment Rules
- Persistence: Variables persist across turns (Jupyter-style). You can directly use the variables defined in the previous turns.
- Output: Use print() to display results you need to see. Due to the limitation on the length of your context, consider the length of printed content and only print what is necessary.
- Files: If a file path is provided, it is in the current directory. Do not write new files.
- Do not write anything into local files.

## Answer Format
If you have sufficient information to solve the user's problem, terminate the tool invocation and wrap your answer with <answer> and </answer>.
- For short-answer questions, directly provide the short answer inside.
- For report-generation questions, wrap the report content inside <answer> and </answer>. Do not write your report in code.
\end{promptbox}

\paragraph{(iii) Specialized data agents.}
For DataMind~\cite{datamind} and DeepAnalyze~\cite{deepanalyze}, we use the
official model checkpoints and public inference scaffolds, keeping their native
prompting and code-execution conventions. We only adapt file paths and answer
extraction to match our benchmark harness. DeepAnalyze is run with a 30-round
code-execution budget; generated \texttt{<Code>} blocks are executed and returned
as \texttt{<Execute>} observations until the model emits \texttt{<Answer>} or the
round budget is exhausted.

\subsection{Grounding Rate Evaluation}
\label{app:traceability-eval}

To quantify the second evaluation axis used in Section~\ref{sec:exp-setup}, we define the \textbf{Grounding Rate (GR)} as claim-level support recoverability from the evidence artifact exposed by a method.

For each sample \(x_i\), we first take the model's final output text \(a_i\) and decompose it into a set of atomic factual claims using an independent LLM call:
\[
\mathcal{C}_i=\{c_{i1},c_{i2},\dots,c_{i|\mathcal{C}_i|}\}.
\]
The decomposition prompt requires the LLM to split the answer into minimal standalone factual claims, preserve the original meaning, avoid introducing new facts, and remove purely rhetorical or hedging text. If the answer is empty or contains no factual content, the claim set is empty. Each atomic claim is therefore treated as a minimal, semantically self-contained factual statement, and the decomposition process does not introduce information beyond the final answer.

For each atomic claim \(c_{ij}\), we then construct the evidence artifact exposed by the method:
\begin{itemize}
  \item \textbf{CodeAct}: the complete solving trajectory is used as the evidence context, i.e., \texttt{thought + code + observation};
  \item \textbf{VeriGraph}: the final evidence subgraph \(\mathcal{G}^*\) is used as the evidence context, concretely realized as the set of claim nodes reachable from the terminal claim node along ancestor relations.
\end{itemize}
This matches the user-facing audit interface of each method: a linear agent exposes a trajectory, whereas \ours exposes a terminal evidence subgraph.

Given the evidence context, we represent its contents as a set of evidence units. To retrieve candidate evidence for each claim, we score evidence units using three complementary heuristics:
\begin{enumerate}
  \item numeric overlap;
  \item lexical token overlap;
  \item exact substring match.
\end{enumerate}
For numeric overlap, we first extract numeric expressions from both the atomic claim and each evidence unit using a separate LLM-based numeric extractor. The extractor is constrained to return only numeric expressions explicitly appearing in the input, including integers, decimals, percentages, currency amounts, ratios, year-like numbers, and signed numbers, without inventing additional values.

For each claim \(c_{ij}\), we keep the top-\(k\) evidence units with the highest heuristic scores, denoted as its candidate evidence set \(\mathcal{E}_{ij}\). The same segmentation, retrieval, and top-\(k\) selection procedure is applied to all methods before judging. An independent LLM judge then determines whether \(\mathcal{E}_{ij}\) is sufficient to support the claim \(c_{ij}\). The judge is given the user question, the atomic claim, and the candidate evidence units, and returns binary labels indicating whether the evidence is relevant to the claim and whether it sufficiently supports the claim. We use the support label to define:
\[
s(c_{ij}) \in \{0,1\},
\]
where \(s(c_{ij})=1\) only if the candidate evidence sufficiently supports the claim, and \(s(c_{ij})=0\) otherwise. The judge is instructed to be conservative: evidence that is missing, unrelated, contradictory, or too weak is treated as unsupported.

The \textbf{Grounding Rate (GR)} is finally defined as:
\[
\mathrm{GR}
=
\frac{\sum_i \sum_{j=1}^{|\mathcal{C}_i|} s(c_{ij})}
{\sum_i |\mathcal{C}_i|}.
\]

GR measures the proportion of atomic claims in the model's final answer that can be explicitly supported by the evidence artifact the method produces. Because the denominator contains only emitted claims, we report GR alongside task scores that measure correctness and completeness. This evaluation separates factual claim decomposition, candidate evidence retrieval, and final support judgment, reducing the chance that unsupported claims are counted as grounded merely due to surface-level overlap. The exact prompts used for atomic claim decomposition, numeric expression extraction, and evidence judging are provided in Section~\ref{app:grounding-rate-prompts}.

%%%%%%%%%%%%%%%%%%%%%%%%%%%%%%%%%%%%%%%%%%%%%%%%%%%%%%%%%%%%%%%%%%%%%%%%%%%%%%%%%%%%%%%%%%%%%%%%%

\subsection{Evaluation Prompts}
The judge prompt for QA accuracy, the rubric prompt for research-task scoring, and the
prompt for our \emph{Grounding Rate} judge are listed in
Listings~\ref{lst:eval-qa-prompt}--\ref{lst:eval-trace-prompt}.

\begin{promptbox}{QA accuracy judge prompt template.}{lst:eval-qa-prompt}
Compare the predicted answer against the golden answer.
Return whether the prediction is correct, allowing equivalent numeric formats and units.
\end{promptbox}

\begin{promptbox}{Grounding Rate judge prompt template.}{lst:eval-trace-prompt}
Given an atomic claim and a candidate evidence set drawn from the method's own evidence context, decide whether the evidence is sufficient to support the claim. Return a binary support decision.
\end{promptbox}

For RL training, the outcome judge receives the task query, task metadata, the golden
answer or official rubric, the terminal answer emitted by the agent, and the serialized
terminal evidence subgraph. The judge assigns an ordinal score
$s \in \{0, 1, \dots, S\}$, where $S$ is the maximum score of the corresponding task
rubric. Exact-answer QA tasks use their binary or normalized accuracy rubric, whereas
open-ended research tasks use the official content / format rubric inherited from the
benchmark.

The rubric checks three aspects in order:
\begin{enumerate}
  \item \textbf{Answer correctness}: whether the final answer matches the golden answer or
        satisfies the task-specific content rubric.
  \item \textbf{Completeness}: whether all requested quantities, comparisons, and caveats
        are covered.
  \item \textbf{Faithfulness to evidence}: whether the terminal answer is supported by the
        submitted evidence graph rather than by unsupported statements outside the graph.
\end{enumerate}
The normalized reward used in Eq.~\ref{eq:outcome_reward} is $s/S$. If terminal extraction
fails or no terminal claim is submitted, the outcome reward is set to zero.

%%%%%%%%%%%%%%%%%%%%%%%%%%%%%%%%%%%%%%%%%%%%%%%%%%%%%%%%%%%%%%%%%%%%%%%%%%%%%%%%%%%%%%%%%%%%%%%%%

\subsection{Grounding Rate Evaluation Prompts}
\label{app:grounding-rate-prompts}

We use three LLM prompts in the Grounding Rate evaluation pipeline: one for decomposing final answers into atomic factual claims, one for extracting numeric expressions used in numeric-overlap retrieval, and one for judging whether the retrieved candidate evidence supports each atomic claim.

\begin{promptbox}{Atomic claim decomposition prompt.}{lst:atomic-claim-prompt}
You decompose an answer into atomic factual claims.

Return strictly valid JSON with this schema:

{
  "claims": ["claim 1", "claim 2"]
}

Rules:
- Split the answer into minimal standalone factual claims.
- Keep the original meaning.
- Do not add new facts.
- Remove purely rhetorical or hedging text.
- If the answer is already one atomic claim, return one item.
- If the answer is empty or contains no factual content, return {"claims": []}.
- Output JSON only.
\end{promptbox}

\begin{promptbox}{Numeric expression extraction prompt.}{lst:number-extract-prompt}
You extract numeric expressions from text.

Return strictly valid JSON with this schema:

{
  "numbers": [
    {
      "text": "original numeric expression as it appears in the input"
    }
  ]
}

Rules:
- Extract only numeric expressions that are part of the answer/claim content.
- Include integers, decimals, percentages, currency amounts, ratios, year-like numbers, and signed numbers.
- Do not invent numbers.
- Preserve the surface form from the input text as closely as possible.
- If there are no numeric expressions, return {"numbers": []}.
- Output JSON only.
\end{promptbox}

\begin{promptbox}{Grounding Rate evidence judge prompt.}{lst:grounding-rate-judge-prompt}
You are a rigorous evidence judge following an ALCE-style claim support protocol.

You will receive:
1. A user question
2. One atomic claim
3. Candidate evidence units from the method-specific context

Your task is to decide:
- supported: whether the provided evidence units contain enough evidence to support the claim
- relevant: whether the provided evidence units are genuinely relevant to the claim rather than mostly unrelated

Return strictly valid JSON:

{
  "supported": 0 or 1,
  "relevant": 0 or 1,
  "reasoning": "brief explanation"
}

Judging rules:
- supported = 1 only if the evidence shown is sufficient for the claim.
- relevant = 1 if the evidence shown is directly about the claim, even if it is not fully sufficient.
- If the evidence is missing, unrelated, contradictory, or too weak, supported = 0.
- Be conservative.
- Output JSON only.
\end{promptbox}

% =====================================================================

\section{Additional Analysis}
\label{app:additional-analysis}

\subsection{Cost and Latency Analysis}
\label{app:cost-latency}

\begin{figure}[htbp]
  \centering
  \includegraphics[width=0.95\linewidth]{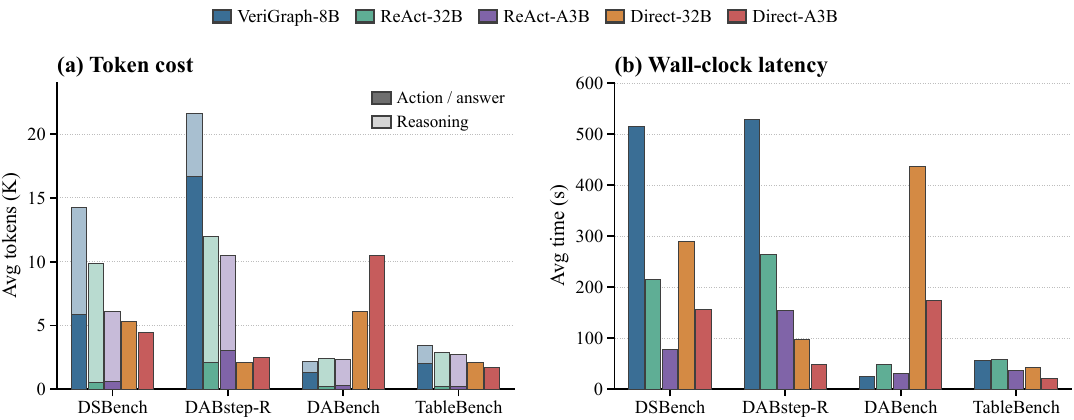}
  \caption{Token cost and wall-clock latency across representative Qwen3-based
  baselines. \textbf{(a)} Average generated tokens, where tool-using systems are
  split into reasoning and code/action tokens and direct systems use a single
  answer-generation stream. \textbf{(b)} Average wall-clock time per example.}
  \label{fig:cost_latency_analysis}
\end{figure}

Figure~\ref{fig:cost_latency_analysis} reports the average token cost and
wall-clock latency of representative Qwen3-based direct and tool-using systems.
For tool-using systems, token counts are separated into reasoning tokens and
code/action tokens; direct-prompting baselines are shown as single-stream answer
generation. The latency measurements include Python-execution overhead for
tool-using systems and evidence-graph materialisation for \ours. As expected,
\ours incurs additional runtime on the long-evidence benchmarks where it builds
larger terminal graphs, while the inference-time cost remains a single rollout
without extra verifier calls.

Beyond the overall cost comparison, Figure 5 also reveals a clear difference in how VeriGraph and ReAct allocate their token budgets. Compared with ReAct, VeriGraph tends to spend a larger fraction of its tokens on code/action generation, whereas ReAct consumes most of its tokens in the reasoning stream. This suggests that VeriGraph effectively shifts part of the reasoning burden from unconstrained natural-language deliberation to executable programs. In other words, operations that require repeated reflection and verification in ReAct can be externalized into code execution in VeriGraph, allowing intermediate computations and evidence aggregation to be performed more deterministically. This design explains why VeriGraph may introduce additional execution latency, but also supports more structured and verifiable reasoning. Overall, the results indicate that VeriGraph trades moderate overhead for explicit computation and graph-based evidence organization, rather than relying primarily on extended internal reasoning.

\subsection{Analysis of Context Management}
\label{app:context-management}

Table~\ref{tab:context-management} compares the baseline runtime with the
context-compressed variant across four representative benchmarks. The main
pattern is that context compression does not necessarily reduce generation
volume: in several settings the code-token budget actually increases, because the policy can afford to emit a more explicit executable trace once earlier history has been compacted. Even so, the wall-clock time is lower on most benchmarks, which is consistent with the reduced effective sequence length seen by the model at each step. In other words, compression shifts computation away from repeatedly reprocessing a long transcript and toward a slightly longer code trace, and the latter is cheaper than carrying the full uncompressed context through every turn.

\begin{table}[htbp]
  \centering
  \small
  \setlength{\tabcolsep}{5pt}
  \renewcommand{\arraystretch}{1.12}
  \caption{Context-management comparison between the baseline runtime and the
  context-compressed variant. Token counts and wall-clock time are rounded to
  whole numbers; aggregate scores are rounded to three decimals.}
  \label{tab:context-management}
  \begin{tabular}{lrrrrrrrr}
    \toprule
    & \multicolumn{4}{c}{\textbf{VeriGraph}} &
      \multicolumn{4}{c}{\textbf{VeriGraph-context}} \\
    \cmidrule(lr){2-5}\cmidrule(lr){6-9}
    Dataset & Thought & Code & Time & Score & Thought & Code & Time & Score \\
    \midrule
    DSBench   & 8431 & 5826  & 515 & 66.43 & 13953 & 5824  & 368 & 65.12 \\
    DABstep-R & 4950 & 16649 & 528 & 3.31 & 5890  & 17328 & 413 & 3.02 \\
    DABench   & 858  & 1316  & 25  & 85.99 & 666   & 1291  & 20  & 84.05 \\
    TableBench& 1460 & 1997  & 56  & 73.58 & 1603  & 2639  & 61  & 72.06 \\
    \bottomrule
  \end{tabular}
\end{table}

The table suggests a clear efficiency trade-off. Across DSBench, DABstep-R, and DABench, the compressed-context variant reduces runtime while maintaining broadly similar aggregate scores, and it does so even when the
number of generated code tokens rises. This indicates that the dominant cost is not token emission itself, but the repeated processing of a long conversational history. Context compression lowers that overhead by shortening the sequence carried into later turns, so the model can spend more of its budget on problem-specific code generation rather than on re-reading prior text. The tablebench result is a mild exception on runtime, but the overall trend remains that context compression improves execution efficiency with only limited impact on the final score.

% \subsection{Analysis of Context Management}
% \label{app:context-management}
% Long-horizon agentic trajectories quickly exhaust the context window because every tool
% call returns observations (table previews, error messages, intermediate dataframes) that
% can be much longer than the model's own action tokens. We provide a quantitative study of
% our fine-grained sliding-window compression strategy described in
% Appendix~\ref{app:verigraph-context}.

% \paragraph{Metrics.} We report (a) \emph{token consumption per turn}, (b) \emph{effective
% turns before truncation}, and (c) \emph{end-task score} for three settings: no compression,
% naive truncation, and our recent-$k$ compression.

% \paragraph{Findings.} \emph{[Figure~\ref{fig:app-context} placeholder.]}
% \dots

% \begin{figure}[htbp]
%       \centering
%       \fbox{\parbox{0.82\linewidth}{\centering Placeholder for context-management analysis:
%       token consumption, effective turns before truncation, and end-task score.}}
%       \caption{Placeholder for the context-management comparison.}
%       \label{fig:app-context}
% \end{figure}

% \subsection{Experience in RL Training}
% 不进行infer/过度infer
% code中发生幻觉的概率比较低

\subsection{Failure Modes in RL Training}
\label{app:rl-failure-modes}

This subsection summarizes the main failure modes we observed when optimizing
the \ours policy with reinforcement learning. Since the reward design is
specified in Appendix~\ref{app:training}, we focus here on practical sources of
instability and on the supervision signals required to make RL reliable for
evidence-graph construction.

\paragraph{Outcome-only RL is too coarse for graph construction.}
Optimizing only the terminal answer reward provides no explicit feedback on the
quality of the intermediate evidence graph. In our experiments, this induces a
length bias: when multiple correct trajectories appear in a group, longer
rollouts receive more total token-level updates because the same
trajectory-level advantage is assigned to every generated token. As a result,
outcome-only RL can favor verbose and redundant derivations, even when a shorter
trajectory is equally correct and yields a cleaner support graph. For graph
construction, additional steps are not inherently beneficial; they enlarge the
failure surface and increase the risk of unsupported claims or unnecessary
detours.

\paragraph{Direct RL from the base model rarely reaches valid terminal states.}
Before the policy learns the atomic graph-writing format, many rollouts fail
before they can be judged: the model may omit required primitives, violate the
output schema, or terminate without producing a submittable final claim set. In
this regime, the final judge is rarely invoked, making the reward extremely
sparse and largely uninformative. We therefore use a cold-start stage with
distilled graph-augmented trajectories, which gives the policy sufficient
syntactic and operational competence to explore the intended RL objective rather
than spending most updates on format violations.

\paragraph{Executability can mask weak semantic transitions.}
The process reward provides the first dense signal by checking whether executed
actions are valid. This is crucial in early training, where malformed code is a
dominant failure mode. However, executability alone does not ensure that a
trajectory is useful for evidence construction. A rollout can be fully
executable while still containing weak or unsupported semantic transitions.
Edge-level supervision on \texttt{infer} is therefore essential: it
distinguishes locally justified derivations from merely plausible ones and
prevents process correctness from degenerating into graph-level noise.

\paragraph{Reliable training requires layered supervision.}
In practice, the strongest signal comes from combining process, \texttt{infer},
and outcome rewards. The process reward keeps the policy within the executable
action space, the inference reward improves the quality of the support graph,
and the outcome reward anchors optimization to end-task correctness. Our
experience suggests that these signals should be used hierarchically rather than
interchangeably: process supervision makes RL trainable, inference supervision
makes the graph meaningful, and outcome supervision keeps the policy aligned
with the final objective.

\subsection{Cross-Model Consistency Analysis}
\label{app:cross-model-consistency}

\begin{figure}[htbp]
  \centering
  \includegraphics[width=1.05\linewidth]{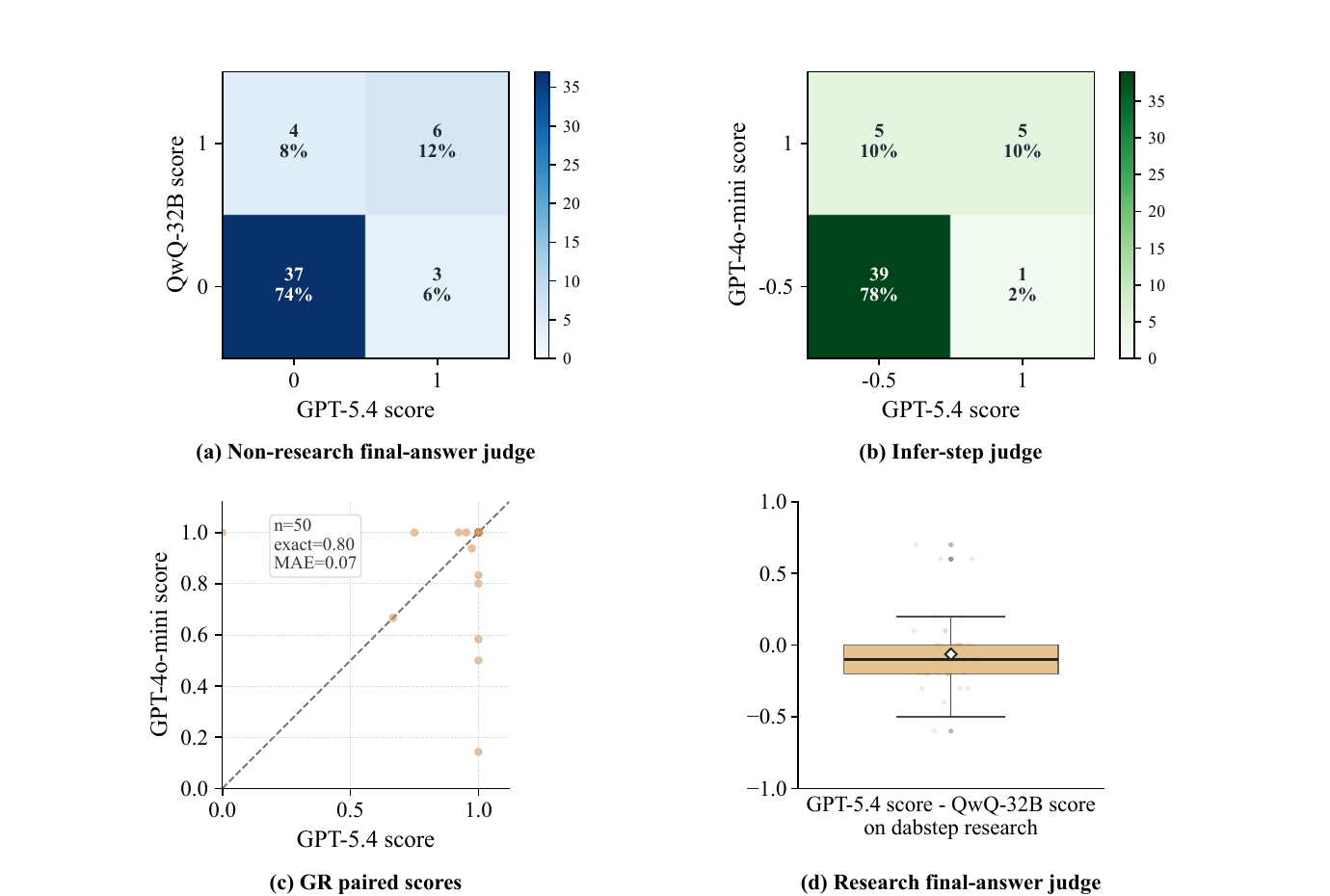}
  \caption{Cross-model consistency analysis. \textbf{(a)} Final-answer judge
  confusion matrix. \textbf{(b)} Infer-step judge confusion matrix.
  \textbf{(c)} Paired grounding recall (GR) scores under two judge backends,
  with exact agreement and mean absolute error reported in the statistics box.
  \textbf{(d)} Signed normalized score difference between GPT-5.4 and QwQ-32B
  on the research subset of LLM-as-judge. For this subset, content and format
  scores are averaged and normalized to the $[0,1]$ range before differencing.}
  \label{fig:consistency-overview}
\end{figure}

To examine whether the observed scores are stable across judge backends, we
compare two different large models on four consistency views: final-answer
judging, infer-step judging, grounding recall (GR), and the research subset of
LLM-as-judge. Figure~\ref{fig:consistency-overview} reports two confusion
matrices for categorical judgments, a paired-score plot for GR, and a normalized
score-difference boxplot for research-style report evaluation. These analyses
are intended as evaluator diagnostics rather than additional task-performance
metrics. For the categorical settings, agreement corresponds to mass on the
diagonal of the confusion matrices. For GR and research-style judging, we
inspect the distribution of paired scores or signed differences to identify
whether the choice of evaluator backend changes the measured signal.

The final-answer judge shows substantial diagonal mass, indicating that the two models usually agree on binary answer correctness, while the off-diagonal cases capture examples where holistic answer quality remains model-sensitive. The
infer-step judge is similarly concentrated on the diagonal, suggesting that local verification of individual inference operations is reproducible across backends. The GR panel provides a score-valued view: most paired citation-recall scores lie close to the identity line, with high exact agreement and low mean absolute error. Finally, the research LLM-as-judge panel compares GPT-5.4 and QwQ-32B after normalizing report-level content and format scores. Its signed differences are centered near zero but have a wider spread than the categorical judges, which is expected because report evaluation uses a finer-grained five-point scale. The final-answer, infer-step, GR, and research comparisons each
use 50 sampled paired examples.

\section{Case Studies}
\label{app:case-study}

To concretely illustrate how \ours operationalizes claim-level verifiability
across qualitatively different analytic regimes, we present two complementary
case studies that bracket the spectrum of tasks targeted by our framework.
The first (Appendix~\ref{app:case-warehouse}) is a \emph{compact,
closed-form} decision-support task in which a single small table admits a
short deterministic derivation and a single recommended action; it is
intended to expose, in fully traceable form, the elementary edge types
(computation, \texttt{bind}, \texttt{infer}) that constitute an evidence
graph. The second (Appendix~\ref{app:case-research-report}) is an
\emph{open-ended, multi-source} research-report task drawn from the
DABstep-research, in which the agent must synthesize dozens of
statistics computed over heterogeneous files into a multi-paragraph
narrative; it is intended to demonstrate that claim-level provenance
remains tractable, and uniformly auditable, as the answer surface and the
underlying reasoning chain grow by an order of magnitude. Together, the
two cases show that the same graph abstraction governs both extremes,
yielding answers whose every numeric assertion can be traced back along a
single edge to the computation node that produced it.

\subsection{Decision Support over a Tabular Source}
\label{app:case-warehouse}
We revisit, in fully expanded form, the warehouse-restocking instance
previewed in Figure~\ref{fig:graph_analysis}(c). The user issues a
decision-support query asking which of three warehouses should be
prioritized for restocking, given a single tabular source listing each
warehouse's current inventory, daily demand, and replenishment lead time
(Table~\ref{tab:case-warehouse-data}). Despite its modest size, the
instance exercises the full \ours pipeline end to end: deterministic
computation over raw cells, \texttt{bind} operations that lift numeric
results into verifiable claims, and a terminal \texttt{infer} step that
composes those claims into the recommended action.

\begin{table}[htbp]
\centering
\caption{Raw table used in the warehouse-restocking case study.}
\label{tab:case-warehouse-data}
\setlength{\tabcolsep}{14pt}
\renewcommand{\arraystretch}{1.15}
\begin{tabular}{@{}cccc@{}}
\toprule
\textbf{Warehouse} & \textbf{Inventory} & \textbf{Daily demand} & \textbf{Lead time} \\
\midrule
A & 120 & 30 & 6 \\
B & 200 & 20 & 4 \\
C & 90  & 15 & 3 \\
\bottomrule
\end{tabular}
\end{table}

The agent first computes three executable quantities for each warehouse. The
\emph{days-to-stockout} value measures how many days current inventory can support demand:
\[
\mathrm{days}_i = \frac{\mathrm{inventory}_i}{\mathrm{daily\_demand}_i}.
\]
The \emph{replenishment gap} compares this quantity with lead time:
\[
\mathrm{gap}_i = \mathrm{days}_i - \mathrm{lead\_time}_i.
\]
Finally, the non-negative shortage-risk score is
\[
\mathrm{risk}_i = \max(0, -\mathrm{gap}_i).
\]
This gives $\mathrm{days}_A=4$, $\mathrm{days}_B=10$, and $\mathrm{days}_C=6$;
$\mathrm{gap}_A=-2$, $\mathrm{gap}_B=6$, and $\mathrm{gap}_C=3$; and risk scores
$\mathrm{risk}_A=2$, $\mathrm{risk}_B=0$, and $\mathrm{risk}_C=0$. Warehouse A is the
only warehouse whose inventory is expected to run out before replenishment arrives.

\begin{figure}[htbp]
  \centering
  \includegraphics[width=0.92\linewidth]{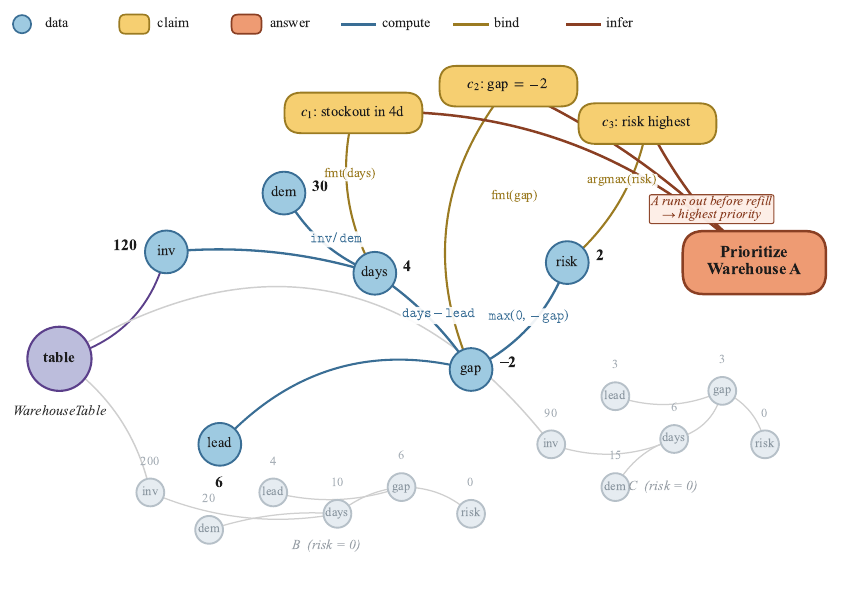}
  \caption{Detailed evidence graph for the warehouse-restocking case. The highlighted
  path is the selected answer chain for Warehouse A; faded nodes show the alternative
  computations for Warehouses B and C. For readability, the figure expands bound runtime
  values into explicit data nodes, while the implemented runtime exports the corresponding
  claim-level DAG.}
  \label{fig:case-graph}
\end{figure}

The evidence graph in Figure~\ref{fig:case-graph} separates the answer into three layers.
First, data-to-data computation edges derive \texttt{A\_days}, \texttt{A\_gap}, and
\texttt{A\_risk} from the raw table fields. Second, \texttt{bind} edges convert selected
computed artifacts into natural-language claims:
\begin{itemize}
  \item $c_1$: Warehouse A will stock out in 4 days.
  \item $c_2$: Warehouse A has a replenishment gap of $-2$ days.
  \item $c_3$: Warehouse A has the highest restocking risk score, with risk 2 versus 0
        for both alternatives.
\end{itemize}
Third, a single \texttt{infer} step combines $c_1$, $c_2$, and $c_3$ into the final claim
$c_4$: Warehouse A should be prioritized for restocking. The faded B/C branches are not
part of the final answer path, but they are important audit evidence: they show that the
agent did not choose A merely because A has low inventory, but because the computed risk
comparison makes A the only positive-risk warehouse.

\begin{promptbox}{Abbreviated VeriGraph trajectory for the warehouse-restocking case.}{lst:case-warehouse-trajectory}
<think>
I need to identify the warehouse with the highest stockout risk. I will compute
days-to-stockout, replenishment gap, and a non-negative shortage-risk score for
each warehouse, then bind the decisive facts and infer the final recommendation.
</think>
<code_interpreter>
warehouse_table = {
    "A": {"inventory": 120, "daily_demand": 30, "lead_time": 6},
    "B": {"inventory": 200, "daily_demand": 20, "lead_time": 4},
    "C": {"inventory": 90, "daily_demand": 15, "lead_time": 3},
}

days = {
    k: v["inventory"] / v["daily_demand"]
    for k, v in warehouse_table.items()
}
gap = {
    k: days[k] - v["lead_time"]
    for k, v in warehouse_table.items()
}
risk = {k: max(0, -gap[k]) for k in warehouse_table}
priority = max(risk, key=risk.get)

c1 = bind(
    "Warehouse A will stock out in {a_days:.0f} days.",
    a_days=days["A"],
)
c2 = bind(
    "Warehouse A has a replenishment gap of {a_gap:.0f} days.",
    a_gap=gap["A"],
)
c3 = bind(
    "Warehouse A has the highest restocking risk score: A={a_risk:.0f}, B={b_risk:.0f}, C={c_risk:.0f}.",
    a_risk=risk["A"], b_risk=risk["B"], c_risk=risk["C"],
)
</code_interpreter>
<tool_response>
Execution Status: Success
Current claims:
[c1] c1 [Context]: Warehouse A will stock out in 4 days.
[c2] c2 [Context]: Warehouse A has a replenishment gap of -2 days.
[c3] c3 [Context]: Warehouse A has the highest restocking risk score: A=2, B=0, C=0.
</tool_response>
<think>
The decisive facts are now claims: A runs out before replenishment and has the
only positive risk score. I can submit the derived recommendation.
</think>
<code_interpreter>
c4 = infer(
    premises=[c1, c2, c3],
    conclusion="Warehouse A should be prioritized for restocking.",
    reasoning=(
        "Warehouse A has only 4 days of inventory against a 6-day lead time, "
        "creating a -2 day replenishment gap, and its risk score of 2 exceeds "
        "the zero risk scores for Warehouses B and C."
    ),
)
submit_answer(c4)
</code_interpreter>
\end{promptbox}

This example illustrates the local auditability targeted by \ours. If the final answer is
challenged, a reviewer can inspect the specific edge where the concern arises: the raw
table extraction, the deterministic computation of \texttt{days}, \texttt{gap}, or
\texttt{risk}, the \texttt{bind} operation that turns a computed value into a claim, or
the \texttt{infer} operation that combines the three claims into the final recommendation.

\subsection{Multi-Source Research Report on a Payments Dataset}
\label{app:case-research-report}
Whereas Appendix~\ref{app:case-warehouse} examined a closed-form
decision over a single tabular source, we now turn to a substantially
more demanding regime that more faithfully reflects practical analyst
workloads. The instance is drawn from the open-ended split of DABstep
and is characterized by an intentionally under-specified prompt, namely
``Create a research report summarizing key observations or notable points
in the data,'' paired with seven heterogeneous source artifacts: two
textual schema descriptions (\texttt{payments-readme.md},
\texttt{manual.md}), a fact table of $138{,}236$ payment transactions
recorded in 2023 (\texttt{payments.csv}), a merchant directory
(\texttt{merchant\_data.json}), a fee-rule catalogue (\texttt{fees.json}),
and two reference dimensions (\texttt{merchant\_category\_codes.csv} and
\texttt{acquirer\_countries.csv}). In contrast to the warehouse case, the
expected deliverable is an open-ended, multi-paragraph narrative rather
than a single scalar answer, and the space of admissible claims is
correspondingly large. This case therefore probes whether claim-level
provenance scales gracefully when the answer surface itself is generative.

\begin{figure}[htbp]
  \centering
  \includegraphics[width=0.95\linewidth]{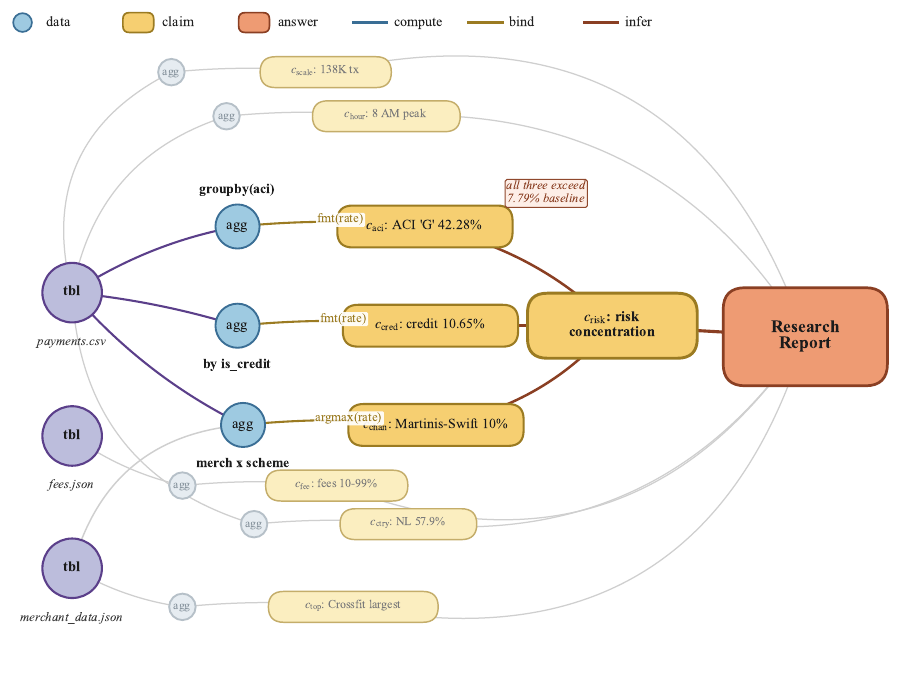}
  \caption{Evidence graph for the payments research-report case. Three source
  tables (left) feed deterministic group-by computations, and each computation is
  lifted into a verifiable claim by a \texttt{bind} edge. The highlighted
  hero subgraph shows a fan-in \texttt{infer} step that derives an
  intermediate claim $c_{\mathrm{risk}}$ (drawn with a thicker border to
  mark it as derived rather than directly bound) from three independent
  binds (ACI breakdown, credit versus non-credit, and merchant$\times$card-scheme),
  all of which exceed the $7.79\%$ dataset-wide dispute baseline. Faded
  satellite claims (e.g., hourly peak, country share, and fee distribution)
  feed the final \texttt{Research Report} answer node directly. Every
  sentence in the report can be traced backwards along a single bind edge
  to the computation node that produced its numeric value.}
  \label{fig:case-research-graph}
\end{figure}

This setting exposes two failure modes that prose-only agents commonly suffer
from. First, with dozens of statistics flowing into a single document, it is
easy for the writer to introduce a number that no upstream computation
actually produced, or to silently re-use a stale value after a later
recomputation, and the reader has no efficient way to localize either error.
Second, the agent must coordinate across multiple files and group-bys (by
merchant, card scheme, hour, ACI code, country, and device), and the chain
of pandas operations that justifies any single sentence is typically long.

\paragraph{What the evidence graph contains.}
Over $29$ reasoning turns, the policy first issues schema-discovery code,
then deterministic aggregation code that produces one statistic per group-by,
and finally a sequence of \texttt{bind} edges that lift each computed scalar
into a natural-language claim. The resulting graph contains $33$ final claims,
an abbreviated list of which is reproduced in
Listing~\ref{lst:case-research-report-claims}. Because every claim is
anchored to the exact computation node that produced its numeric value, the
report is no longer a wall of statistics but rather a view over a checkable
graph. Table~\ref{tab:case-research-claims} shows representative claim
families together with the upstream computations they bind to.

\paragraph{Trajectory shape.}
Listing~\ref{lst:case-research-trajectory} shows an abbreviated slice of the
trajectory. A deterministic group-by produces the per-ACI dispute rate, a
single \texttt{bind} step lifts the salient row into a verifiable claim, and
a subsequent fan-in \texttt{infer} step composes a ``risk-concentration''
claim from the ACI, credit-card, and merchant-channel binds. The same
pattern is repeated for each claim family, and the final report node is
then an \texttt{infer} over the selected claim subset.

\begin{promptbox}{Abbreviated VeriGraph trajectory for the payments research-report case.}{lst:case-research-trajectory}
<think>
The request is open-ended. I will materialise one statistic per group-by,
bind each into a claim, and only then synthesise the report so that every
sentence is traceable to a specific computation.
</think>
<code_interpreter>
import pandas as pd, json
payments = pd.read_csv("payments.csv")
fees     = pd.DataFrame(json.load(open("fees.json")))
merchants = pd.DataFrame(json.load(open("merchant_data.json")))

# --- per-ACI dispute rate ---
aci_stats = (
    payments.groupby("aci")["has_fraudulent_dispute"]
            .agg(["mean", "count"])
            .sort_values("mean", ascending=False)
)
top_aci          = aci_stats.index[0]
top_aci_rate     = aci_stats.loc[top_aci, "mean"]
top_aci_count    = int(aci_stats.loc[top_aci, "count"])
other_aci_rate   = aci_stats.drop(top_aci)["mean"].max()

c_aci = bind(
    "ACI code '{aci}' has a dispute rate of {rate:.2%} over {n} transactions, "
    "compared to a max of {other:.2%} for the remaining ACI codes.",
    aci=top_aci, rate=top_aci_rate, n=top_aci_count, other=other_aci_rate,
)
</code_interpreter>
<tool_response>
Execution Status: Success
Current claims:
[c_aci] ACI code 'G' has a dispute rate of 42.28% over 25463 transactions,
        compared to a max of 0.00% for the remaining ACI codes.
</tool_response>
<think>
I now repeat the same pattern for the credit-card flag, the
merchant$\times$card-scheme cross-tab, the hourly profile, the fee
catalogue, and the per-country split. Each statistic becomes its own bind.
Once all binds are in place, a single infer step composes the report.
</think>
<code_interpreter>
# ... per-merchant, per-hour, per-card-scheme, fees, country binds ...

c_risk = infer(
    premises=[c_aci, c_credit, c_merchant_channel],
    conclusion=(
        "Fraud risk is concentrated in specific channels: ACI 'G' (42.28%), "
        "credit cards (10.65%), and Martinis_Fine_Steakhouse-SwiftCharge (10%)."
    ),
    reasoning=(
        "All three binds independently identify channels whose dispute rate "
        "exceeds the 7.79% dataset-wide baseline by a large margin."
    ),
)

c_report = infer(
    premises=[c_scale, c_overall, c_aci, c_merchant_top, c_hour,
              c_country, c_fee, c_credit, c_risk, ...],
    conclusion="<multi-paragraph report bound to the above claims>",
)
submit_answer(c_report)
</code_interpreter>
\end{promptbox}

\begin{table}[htbp]
\centering
\caption{Representative claim families in the payments research report and
the deterministic computations they are bound to. Each row corresponds to a
distinct subgraph in the evidence graph, and the final report is an
\texttt{infer} step over a selected subset of these claims.}
\label{tab:case-research-claims}
\footnotesize
\begin{tabular}{@{}p{0.30\linewidth}p{0.32\linewidth}p{0.30\linewidth}@{}}
\toprule
\textbf{Claim family} & \textbf{Upstream computation} & \textbf{Example bound claim} \\
\midrule
Dataset scale & \texttt{len(payments)}, \texttt{nunique(merchant)}, \texttt{nunique(card\_scheme)} & 138{,}236 transactions across 5 merchants and 4 card schemes. \\
Overall fraud rate & \texttt{mean(has\_fraudulent\_dispute)} & Overall fraudulent-dispute rate is $7.79\%$. \\
ACI breakdown & \texttt{groupby(aci).agg(rate, count)} & ACI code \texttt{G}: $42.28\%$ dispute rate over $25{,}463$ transactions, vs.\ $0\%$ for other ACI codes. \\
Per-merchant ranking & \texttt{groupby(merchant).agg(...)} & \texttt{Crossfit\_Hanna} is the largest merchant ($55{,}139$ transactions, mean $92.07$\,EUR). \\
Hourly fraud profile & \texttt{groupby(hour\_of\_day).mean(...)} & Hour $8$ has the highest dispute rate ($6.43\%$) within the 8--10\,AM window. \\
Channel risk concentration & join over \texttt{aci}, \texttt{is\_credit}, \texttt{merchant}$\times$\texttt{card\_scheme} & Risk is concentrated in ACI \texttt{G} ($42.28\%$), credit cards ($10.65\%$), and \texttt{Martinis\_Fine\_Steakhouse}-\texttt{SwiftCharge} ($10.00\%$). \\
Fee structure & \texttt{describe(rate, fixed\_amount)} on \texttt{fees.json} & Fee rates span $10\%$--$99\%$ (mean $54.3\%$); fixed amounts $0$--$0.14$\,EUR (mean $0.069$\,EUR). \\
Geographic concentration & \texttt{groupby(acquirer\_country).size()} & Netherlands accounts for $57.9\%$ of acquired volume. \\
\bottomrule
\end{tabular}
\end{table}

\paragraph{Auditability under realistic length.}
This case makes concrete why the evidence graph matters more, not less, as
the answer becomes longer. The final report contains roughly twenty
distinct numeric claims spread over several paragraphs. Without per-claim
provenance, a reviewer who suspects, for instance, the $42.28\%$ figure
for ACI \texttt{G} would have to re-derive it from the raw trajectory log.
With \ours, the same reviewer follows a single edge from the sentence to
its bind, and then from the bind to the deterministic
\texttt{groupby(aci)} computation that produced it. The same property
applies to fan-in claims such as the risk-concentration sentence: the
reviewer can verify each premise independently and check that the
\texttt{infer} step does not introduce an unsupported numeric value.

A second observation is that the graph cleanly exposes redundancy and
mild inconsistency that prose alone would hide. Several binds in this
case, such as two slightly different statements of the same hourly fraud
peak, map to overlapping computation nodes, so that the deduplication
needed to produce a clean report becomes a graph operation rather than a
re-reading of free text. We view this as a representative example of the
kinds of audits that become tractable once the answer is a graph.

\begin{promptbox}{Final claim set produced for the payments research-report case (abbreviated).}{lst:case-research-report-claims}
[c01]  138236 transactions in 2023, 5 merchants, 4 card schemes.
[c02]  Overall fraudulent-dispute rate: 7.79%.
[c03]  ACI 'G': 42.28% dispute rate over 25463 transactions; other ACI codes ~0%.
[c04]  Crossfit_Hanna is the largest merchant: 55139 transactions, mean 92.07 EUR.
[c05]  Per-merchant dispute rates are tightly clustered (7.68%--8.00%).
[c06]  Acquirer countries: 5 distinct; Netherlands dominates with 57.9%.
[c07]  Hour 8 is the peak transaction hour; highest hourly dispute rate in 8--10 AM.
[c08]  Card schemes: GlobalCard / NexPay / TransactPlus / SwiftCharge, 13733--48150 tx.
[c09]  Fee rates: min 10%, max 99%, mean 54.3%; fixed amounts 0.00--0.14 EUR.
[c10]  Device mix: 'Other' (22.4%) and Android most frequent.
[c11]  Credit-card dispute rate: 10.65% vs. 0.00% non-credit.
[c12]  49 distinct merchant category codes (3000--9399).
[c13]  Highest merchant-card-scheme dispute rate: Martinis_Fine_Steakhouse-SwiftCharge (10%).
[c14]  Lowest merchant-card-scheme dispute rate: Belles_cookbook_store-GlobalCard (7.11%).
[c15]  Total transaction value: 12,697,297.46 EUR; mean 91.85 EUR.
... (further binds for fee distribution, MCC ranges, and merchant directory metadata)
[c33]  Risk-concentration infer: ACI 'G', credit cards, and Martinis-SwiftCharge
       jointly account for the channels whose dispute rate exceeds the 7.79% baseline.
\end{promptbox}

\section{Reproducibility and Compliance}
\label{app:reproducibility}

\subsection{Compute Resources}
\label{app:compute}
All training and evaluation runs were performed on a cluster of
A100-80G GPUs.

\begin{itemize}
  \item \textbf{SFT}: $4 \times$ A100-80G GPUs, bfloat16, DeepSpeed ZeRO-3, approximate
        wall-clock time $24$ hours.
  \item \textbf{RL}: $8 \times$ A100-80G GPUs, with rollout generation and policy
        updates co-located on the same nodes; approximate wall-clock time $48$ hours.
  \item \textbf{Inference / evaluation}: $4 \times$ A100-80G GPUs for agent rollouts
        and LLM-as-judge evaluation. Average per-benchmark wall-clock time on \ours-8B
        is approximately $1.5$ hours (TableBench), $0.5$ hours (InfiAgent-DABench),
        $1.0$ hour (DSBench), and $0.8$ hours (DAB-Step Research).
\end{itemize}
Including preliminary runs, ablations, and failed experiments not reported in the main
text, the full research project consumed roughly $1.5\times$ the budget summarized
above.

\subsection{Licenses for Existing Assets}
\label{app:licenses}
We use the following upstream assets in compliance with their respective licenses. All
datasets are used solely for non-commercial research evaluation; we do not redistribute
the upstream data or model weights.

\begin{itemize}
  \item \textbf{TableBench}~\cite{tablellm}: Apache License 2.0.
  \item \textbf{InfiAgent-DABench}~\cite{infiagentdabench}: Apache License 2.0.
  \item \textbf{DSBench}~\cite{dsbench}: MIT License.
  \item \textbf{DABstep}~\cite{dabstep}: Apache License 2.0.
  \item \textbf{Qwen3-8B}~\cite{qwen3} (base model): Apache License 2.0.
  \item \textbf{MS-Swift}~\cite{msswift} (SFT framework): Apache License 2.0.
  \item \textbf{verl}~\cite{verl} (RL framework): Apache License 2.0.
\end{itemize}
Our own code and trained checkpoints will be released under the Apache License 2.0,
consistent with the licenses of the upstream training frameworks and base model.

\section{Limitations and Future Directions}
\label{app:limitations}

While \ours achieves strong empirical results, several limitations remain that
also point to promising directions for future work.

\paragraph{Internalizing the evidence graph at scale.}
Constrained by compute budget and the size of the available trajectory data,
our post-training is performed at a relatively small scale. Although this is
already sufficient to deliver consistent gains, we believe that scaling up
post-training, or injecting structured evidence-graph supervision earlier in
the pre-training stage, would let the model internalize graph construction as
a native reasoning skill rather than a learned interface, further improving
both the quality and the efficiency of the resulting reasoning.

\paragraph{General task coverage.}
Our experiments and evaluation focus on data-intensive scenarios, where
traceability and factuality are particularly important and currently
under-served. As agent capabilities continue to mature and real-world
deployments expand, we expect attention to shift from raw accuracy toward
auditability and faithfulness across many more task families. Extending
\ours to multimodal reasoning, cross-document synthesis, and long-horizon
tasks that span extended time frames would expose new structural challenges
(e.g., visual evidence, retrieval provenance, temporal dependencies) and
new opportunities for evidence-graph supervision.

\paragraph{Richer reinforcement-learning signals.}
The composite reward we introduce already yields meaningful improvements,
but its current realisation is intentionally conservative: it builds on
relatively simple graph-structural features and edge-level verifier checks.
A natural next step is to evaluate the entire evidence graph with more
expressive models, for example graph neural networks that score global
coherence and informativeness, or learned critics calibrated against audited
reasoning errors. Incorporating targeted human feedback into the reward
design is another promising avenue, which could guide the policy toward
constructing higher-quality, more genuinely useful evidence graphs.

\section{Statement on the Use of LLMs}
During the preparation of this manuscript, we use LLMs as a general-purpose assistance tool. The primary role of the LLM is to aid in improving the clarity and readability of the text, as well as to accelerate the implementation of our research ideas. Specific applications include: (1) Language and Grammar Correction: Polishing sentence structure, correcting grammatical erros, and refining word choices to enhance the overall quality of the writing. (2) Paraphrasing and Style Refinement: Rephrasing sentences and paragraphs to ensure consistency in tone and style throughout the paper. (3) Code Implementation Assistance: Generating code snippets and providing debugging support to help implement the proposed algorithms and experimental setups.

It should be noted that all core research concepts, experimental design, data analysis, and conclusions are developed exclusively by the human authors. Any content or suggestions generated by the LLM, including code, are critically checked, and substantially edited by the authors to ensure accuracy. The authors take full responsibility for the final content of this paper.

\end{document}